\documentclass[10pt,conference]{IEEEtran}
\IEEEoverridecommandlockouts

\usepackage{xspace} 
\usepackage{hyperref}
\usepackage{multirow}
\usepackage{colortbl}
\usepackage{algorithm}
\usepackage{booktabs}
\usepackage{algpseudocode}
\usepackage{cite}
\usepackage{amsmath,amssymb,amsfonts}
\usepackage{graphicx}
\usepackage{textcomp}
\usepackage{xcolor}
\def\BibTeX{{\rm B\kern-.05em{\sc i\kern-.025em b}\kern-.08em
    T\kern-.1667em\lower.7ex\hbox{E}\kern-.125emX}}

\usepackage{tikz}
\usetikzlibrary{positioning,calc}

\usepackage{amssymb}
\usepackage{pifont}

\tikzset{
  box/.style={
    draw,
    rounded corners,
    minimum height=8mm,
    text width=3.2cm,   
    align=center
  }
}

\definecolor{navyblue}{RGB}{0, 0, 128}
\newcommand*\circled[1]{\tikz[baseline=(char.base)]{
            \node[shape=circle, fill=navyblue, text=white, inner sep=1pt] (char) {#1};}}

\newcommand{\name}{{\tt MM-BEV}}

\begin{document}

\title{MM-BEV: Enhancing Timeliness by Computing Where and When it Matters}

\author{
\IEEEauthorblockN{
Liangkai~Liu\IEEEauthorrefmark{1} and
Kang~G.~Shin\IEEEauthorrefmark{2}
}
\IEEEauthorblockA{
\IEEEauthorrefmark{1}Department of Computer Science, Texas Tech University, USA\\
\IEEEauthorrefmark{2}Department of Computer Science and Engineering, University of Michigan, USA
}
}


\maketitle

\begin{abstract}
Multi-modality Bird's-Eye-View (MM-BEV) perception, which combines
LiDAR's depth precision with cameras' dense semantics, is essential
for autonomous vehicles (AVs) but difficult to run in real time 
due to its high compute cost and non-ideal sensing conditions.
State-of-the-art (SOTA) approaches compress single-modality detectors 
and fail to account for three key aspects: (1) structured 
\emph{intra-modality sparsity} of camera and LiDAR inputs, 
(2) the \emph{inter-modal timing misalignment} between
them, and (3) most detected objects are not relevant to
the planner's immediate control action.

We present \name{}, a real-time MM-BEV system organized around 
a fundamental principle: \emph{compute where and when it matters}. 
\name{} decomposes perception into a \emph{mandatory} 
part---safety-critical objects within the ego's braking distance 
and short time-to-collision (TTC)---and an \emph{optional} part, 
prioritizing the mandatory part for fast completion while 
down-sampling or shedding optional work if the compute budget is tight. 
It combines four mechanisms: (i) a \emph{criticality-ranked 
temporal-ROI selector} driven by motion-extrapolated 
previous-frame detections;
(ii) \emph{sparse ROI-aware feature extraction} with a shared-shape
camera crop at context-adaptive resolution and ROI-aware LiDAR
voxelization; (iii) a \emph{latency-aware coordinator} that schedules
LiDAR sweeps, image resolution, and keyframes from scene dynamics and
TTC; and (iv) an \emph{asynchronous scheduler} that decouples sensing
from inference and skips stale frames to keep end-to-end (e2e) latency low.

On the nuScenes dataset, \name{} reduces inference
latency by $1.96\times$ and e2e latency by $2.93\times$,
preserving \emph{geometry-critical} recall exactly and dropping
\emph{safety-critical} recall by only $0.2$\,pp. On a Clearpath Husky
A300 robot with an Ouster-128 LiDAR, BEV cameras, and a Jetson AGX
Orin, \name{} achieves a further $2.11\times$ mean-latency reduction,
corroborating its potential for real-world autonomy.

\end{abstract}



\section{Introduction}


An autonomous vehicle (AV) must perceive its surroundings in real
time---planner cannot use late sensor frames.
Bird's-Eye-View (BEV) perception has become the dominant
representation for this task: by projecting onboard camera
and LiDAR data into a unified 360\textdegree{} top-down view
\cite{ma2023visioncentric,BEV-Perception-Survey-2024}, it blends
LiDAR's precise depth with the cameras' dense semantic cues
for robust detection across diverse
conditions~\cite{liu2023bevfusion,harley2023simple}.
However, on a resource-limited platform aboard AVs, accuracy and
the real-time constraint pull against each other:
fusing two high-rate, high-resolution sensor streams is expensive,
and multi-modality BEV perception is notoriously hard to run in real
time \cite{liu2023bevfusion,bai2022transfusion,yan2023cross,
huang2024detecting,hu2023planning,liu2024rt, liu2022prophet}.

Prior work has approached this real-time/latency challenge along 
three axes, each of which captures only a slice of the problem.
The standard solution is to make the detector cheaper---compressing
dense models into sparse ones~\cite{liu2023sparsebev,
xie2023sparsefusion, yan2023cross,ren2018sbnet, chen2024m}---but
this still treats every region of the input uniformly. 
Region-of-Interest (ROI) heuristics come closer, yet have been 
explored almost entirely on camera images and \emph{within} a single
modality~\cite{zhang2021elf,jiang2021flexible}; they never allocate
compute \emph{across} camera and LiDAR explicitly. On the timing
axis, most pipelines assume pre-synchronized inputs or absorb
worst-case delays without exploiting the detection task's
structure~\cite{li2023worst,sun2023seam}. The two efforts closest to
ours each capture only one slice: RT-BEV~\cite{liu2024rt} couples
ROI-based detection with synchronization but is camera-only and
ignores cross-modal redundancy and asynchronous fusion;
FLEX~\cite{xu2024flex} adapts the LiDAR sweep count per view under a
global time budget, but at coarse per-view granularity and without
exploiting frame-to-frame temporal coherence. Moreover, every
one of these systems is judged by all-object accuracy---the metric
that mislabels a planning-safe speedup as an accuracy loss.

To understand where on-board compute is actually redundant, we
profiled the multi-modality BEV pipeline (\S\ref{sec:empirical-study}).
The inefficiency is \emph{structured}, with the two modalities
incurring redundant work in \emph{complementary} dimensions. Cameras are
\emph{spatially sparse}: only a small, stable fraction of image regions
ever contains an object, yet a dense backbone pays for every pixel.
LiDAR is \emph{temporally redundant}: aggregating more sweeps yields
quickly diminishing the accuracy gain while latency grows steadily.
Compounding both, the sensors are \emph{asynchronous}---running at
different rates with non-negligible communication delays, so their
timestamps are rarely aligned and the inter-modal offsets silently 
degrade fusion~\cite{song2024timealign,li2020towards}.

These findings frame our research problem. Existing BEV perception 
systems ignore this structure---they spend compute uniformly across 
input they could route, and across sweeps they could shed---and they 
are evaluated by a metric that hides whether the spared compute would
have mattered for the planner. For a real-time safety system, the metric
is wrong: as in \emph{imprecise computation}, the perception task
has a \emph{mandatory} part---the \emph{safety-critical} objects 
within the ego's braking distance and short time-to-collision (TTC), 
which the planner must act on within TTC---and an \emph{optional} part. 
The right question is, therefore, not ``how fast can the model be,'' 
but \emph{which work is mandatory, and how do we prioritize it 
for fast completion?}

We answer this questionby developing \name{}, a real-time 
multi-modality BEV perception system organized around a
fundamental principle: compute \emph{where and when it matters}. 
\name{} routes both modalities' compute by the previous frame's
motion-extrapolated detections (\emph{temporal-ROI} routing), 
and co-designs per-frame workload reduction with pipeline 
scheduling so that the mandatory detections are
always processed first and with the lowest possible latency.

\name{} consists of four parts.
\emph{(i) A criticality-ranked temporal-ROI selector} projects the
previous frame's motion-padded detections into per-modality ROIs and
ranks them by geometry and safety criticality, so the mandatory set is
always covered.
\emph{(ii) Sparse ROI-aware feature extraction} confines the camera
backbone to a shared-shape ROI crop at a context-adaptive resolution,
and confines LiDAR voxelization to the ROI point set.
\emph{(iii) A latency-aware coordinator} holds the per-frame budget by
scheduling the LiDAR sweep window, image resolution, and periodic
full-scene keyframes from scene dynamics and TTC.
\emph{(iv) An asynchronous streaming scheduler} decouples sensing from
inference and processes only the freshest synchronized sensor data, so
the pipeline stays responsive instead of falling behind on a growing
queue.

We implement \name{} and evaluate it e2e
in ROS on the nuScenes
dataset~\cite{caesar2020nuscenes,liu2023bevfusion,quigley2009ros} on a
single NVIDIA RTX\,3080 GPU. Temporal-ROI routing with adaptive
resolution cuts mean inference latency $1.96\times$ ($202\!\to\!103$\,ms)
for the LiDAR+camera detector and $1.93\times$ ($136\!\to\!70$\,ms)
for the LiDAR-only detector; the asynchronous scheduler cuts
e2e latency $2.93\times$ ($1006\!\to\!344$\,ms) and raises
processed frames per scene from $\sim$$10$ to $60$. Most importantly, this
does not cost the mandatory work: geometry-critical recall is preserved \emph{exactly} and
safety-critical recall drops only $0.2$ points. A deployment on a
Clearpath Husky\,A300 with a Jetson AGX Orin reaches a further
$2.11\times$ mean-latency reduction, confirming the gains transfer
to real-world autonomous systems.

This paper makes four contributions:

\begin{itemize}
    \item \textbf{Criticality-aware perception.} A mandatory/optional
    decomposition of BEV detection from ego state, and a
    criticality-bucketed metric that scores latency--accuracy
    trade-offs against the objects a planner must act on.

    \item \textbf{Cross-modal temporal-ROI routing.} A
    previous-frame-driven selector that focuses both modalities'
    compute---shared-shape camera ROI, motion-padded LiDAR ROI,
    adaptive sweeps and resolution---prioritized so the mandatory set
    is always covered.

    \item \textbf{A latency-aware coordinator with asynchronous
    scheduling} that keeps e2e latency low by always running
    inference on the freshest synchronized sensor data and adapting
    the keyframe schedule, LiDAR sweep window, and image resolution
    to scene dynamics.

    \item \textbf{An e2e evaluation} on nuScenes in ROS quantifying
    which sparsity
    levers pay off, plus a real-world validation on a Husky\,A300 with
    a Jetson AGX Orin.
\end{itemize}

\section{BEV Perception Pipeline}
\label{sec:background}


State-of-the-art (SOTA) Bird’s-Eye View (BEV) perception 
typically relies on indiscriminate processing of large-scale input data 
from both LiDARs and cameras~\cite{BEV-Perception-Survey-2024,
liu2023bevfusion,yan2023cross,huang2024detecting,liang2022bevfusion,
bai2022transfusion,xie2023sparsefusion}. 

\begin{figure*}[ht]
    \centering
    \includegraphics[width=.8\textwidth]{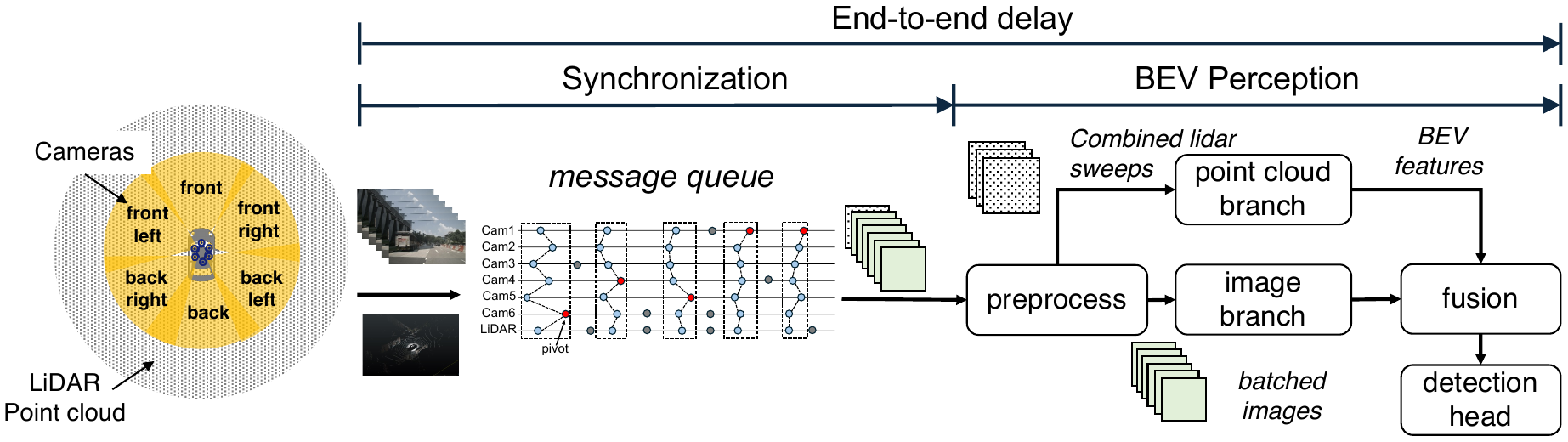}
    \vspace{-1mm}
    
    \caption{End-to-end pipeline for BEV perception.}
    \label{fig:bev-pipeline}
\end{figure*}

Fig.~\ref{fig:bev-pipeline} present a detailed BEV perception 
pipeline with both LiDAR and surrounding cameras. 
The pipeline usually consists of two critical stages: communication 
and detection~\cite{liu2024rt}. The communication stage involves two 
subprocesses: capturing and synchronization. In the former, 
multiple cameras (typically including front, front-right, 
front-left, back, back-left, and back-right) and LiDAR sensors 
independently acquire environmental data~\cite{caesar2020nuscenes,
sun2020scalability,geyer2020a2d2,wilson2023argoverse,houston2021one}. 
The latter (synchronization) ensures temporal alignment across these 
data streams by matching timestamps associated with each sensor's 
data, maintaining the coherence required for precise perception. 
Once synchronized, the detection stage begins by processing the 
sensor data via two parallel branches—one for camera-based images 
and another for LiDAR point clouds. Multiple lidar sweeps are usually 
accumulated for a richer point cloud as input~\cite{caesar2020nuscenes}. 
Each branch independently extracts meaningful spatial features from its 
respective sensor modality. These separate features are then converted 
into a unified BEV representation, creating aligned BEV feature maps. 
A detection head is then employed on these fused BEV features to 
effectively detect and localize relevant objects in the environment, 
ultimately outputting accurate and timely detection results essential 
for AV operation.

\section{Empirical Study}
\label{sec:empirical-study}

We derive \name's design principles from empirical studies
of the SOTA BEV perception pipeline with LiDAR and camera sensors.
We identify three key insights for accuracy and real-time capability
in BEV perception with LiDAR and cameras: 1) fine-grained ROIs are
necessary; 2) asynchronous sensor timing leads to misalignment; and
3) most detected objects are safety-noncritical, so accuracy must be
measured against the safety-relevant subset.

\subsection{Why Fine-Grained ROIs?}

\begin{figure}[h]
    \centering
    \includegraphics[width=.85\columnwidth]{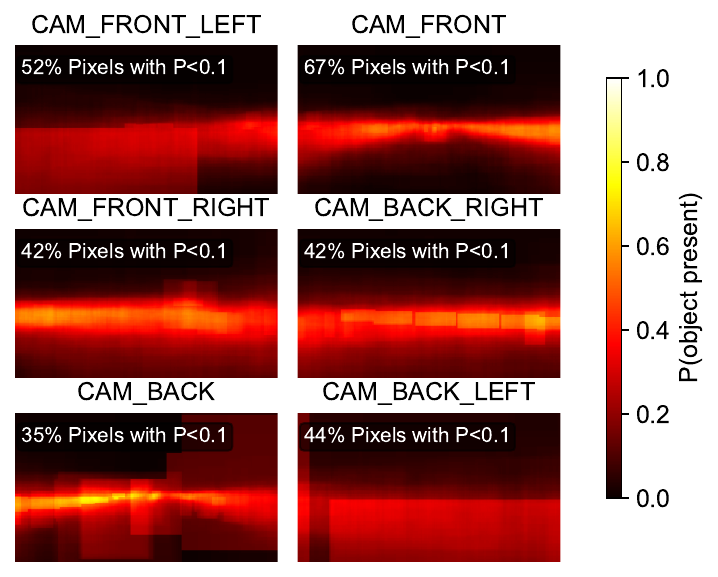}
    \vspace{-2mm}
    \caption{Object-presence probability heatmaps across the six
    nuScenes cameras. Dark areas indicate regions with low per-region
    probability $\bar{P}$; percentages (if shown) denote the share of
    regions with $\bar{P}<0.1$.}
    \vspace{-2mm}
    \label{fig:probability-heatmap-nuscenes}
\end{figure}

\noindent \textbf{Camera inputs are spatially sparse.}
Surround-view camera inputs used for BEV exhibit strong
\emph{spatial sparsity}: most pixels (or image regions) rarely overlap
with any object, while informative signal concentrates in a small,
stable subset of regions (e.g., road lanes, horizon near traffic
flow).
To quantify this, we compute a per-pixel \emph{object-presence
probability} for each camera using nuScenes keyframes. For every
frame, all annotated 3D boxes are projected to the image plane via
calibrated intrinsics/extrinsics into a binary mask (1 if inside any
projected box, 0 otherwise). Summing over $N$ frames and normalizing
yields:
\[
P(x,y)=\frac{1}{N}\sum_{t=1}^{N}\mathbf{1}\!\left[(x,y)\in \bigcup_{k} \mathrm{Proj}\!\left(\mathrm{Box}_{k,t}\right)\right].
\]
We apply a light Gaussian blur to reduce aliasing, and then average
within a regular grid to obtain per-region probabilities $\bar{P}$.
Regions with $\bar{P}<\tau$ (we use $\tau{=}0.1$) are labeled
\emph{low-saliency}. As shown in
Fig.~\ref{fig:probability-heatmap-nuscenes}, large, structured
low-saliency areas persist across views, revealing that coarse,
frame-level pruning cannot exploit where the signal actually lies.
This motivates \emph{fine-grained ROIs} that (i) activate computation
only on high-saliency regions, (ii) reduce fusion and synchronization
costs by aligning far fewer spatial elements, and (iii) preserve
accuracy by retaining a few regions that 
contain objects.

\vspace{1mm}
\noindent \textbf{LiDAR sweeps are temporally redundant.} 
In addition, we observe substantial redundancy when
combining multiple sweeps to enrich the LiDAR's point cloud. 
For LiDAR-only BEV perception, we adjust the 
number of temporal sweeps from 1 to 9 (default in nuScenes) to 
examine how additional point cloud data affects accuracy and 
latency~\cite{caesar2020nuscenes, nuscenes_detection}. 


\begin{figure}[ht]
    \centering
    \includegraphics[width=\columnwidth]{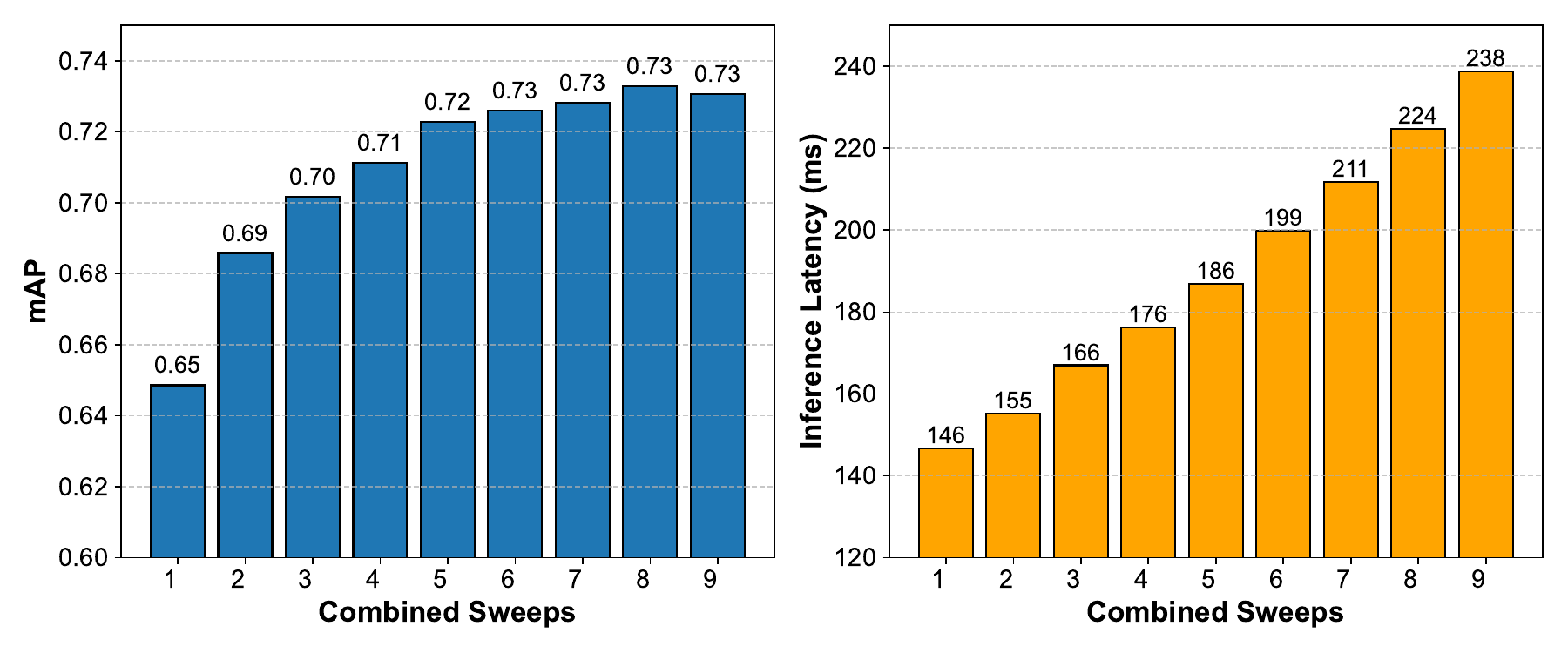}
    \vspace{-4mm}
    \caption{mAP and inference latency with different LiDAR sweeps.}
    \label{fig:lidar-sweeps-map-time}
\end{figure}

Fig.~\ref{fig:lidar-sweeps-map-time} quantifies the trade-off 
between accuracy and latency as a function of the number of 
aggregated LiDAR sweeps. Increasing the number of sweeps from 1 to 9 
results in a steady rise in mean Average Precision (mAP) from 0.65 
to 0.73, but the accuracy gain plateaus beyond 5 sweeps. 
In contrast, inference latency increases almost linearly, from 146\,ms 
with a single sweep to 238\,ms with 9 sweeps. This observation 
reveals temporal redundancy in LiDAR input— additional sweeps 
provide a diminishing return of accuracy while incurring a substantial 
latency overhead. These findings motivate an adaptive sweep 
selection and downsampling mechanism that dynamically balances between sensing richness and 
computational efficiency based on scene motion or 
environmental complexity.



\vspace{0.5em}
\noindent\textbf{Insight 1.} \textit{Spatial sparsity in cameras and
temporal redundancy in LiDAR together call for
\emph{fine-grained, context-aware ROI selection}: activate only the
camera regions (and LiDAR sweeps) that matter, reducing compute and
fusion overheads without sacrificing accuracy.}

\subsection{Why Co-Optimize Sync and Detection?}

\noindent \textbf{Asynchronous sensor timing is the norm, not an exception.}
Most BEV perception pipelines assume perfectly synchronized LiDAR and camera 
inputs~\cite{caesar2020nuscenes, sun2020scalability}, yet real systems run sensors 
on separate clocks and at different rates, introducing unavoidable temporal 
offsets~\cite{xia2025robust,song2024timealign,liu2024rt}. 
In nuScenes, LiDAR samples at $20$\,Hz while the surround cameras run at $12$\,Hz; 
the official dataset exposes only $2$\,Hz \emph{keyframes} for training/evaluation. 
Even when an \texttt{ApproximateTime} policy in \texttt{message\_filters} is used 
for synchronization, residual offsets on the order of $50$--$125$\,ms are common 
due to sampling-rate mismatch and bus jitter. 
When multi-sweep LiDAR aggregation is enabled to densify the point cloud, the effective 
sensing window can span up to $\sim\!500$\,ms (keyframe at $2$\,Hz + several 
preceding non-keyframe sweeps), further widening inter-modal misalignment.

\begin{figure}[ht]
    \centering
    \includegraphics[width=\columnwidth]{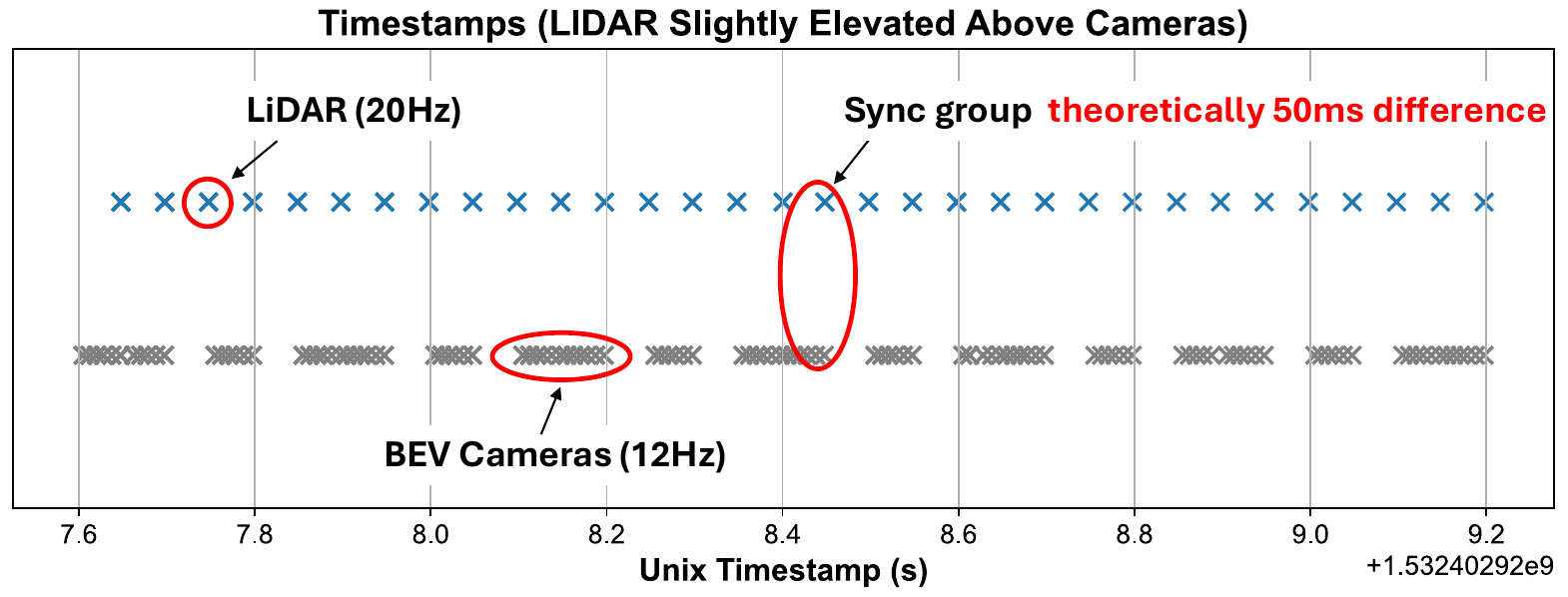}
    \vspace{-4mm}
    \caption{Timestamp distributions of LiDAR (20\,Hz) and BEV cameras 
     (12\,Hz) in nuScenes. Sampling-rate mismatch and asynchronous 
     publication create inherent offsets ($\gtrsim\!50$\,ms) even 
     before bus/jitter effects.}
    \label{fig:sync-time-diff}
\end{figure}

\begin{figure}[ht]
    \centering
    \includegraphics[width=\columnwidth]{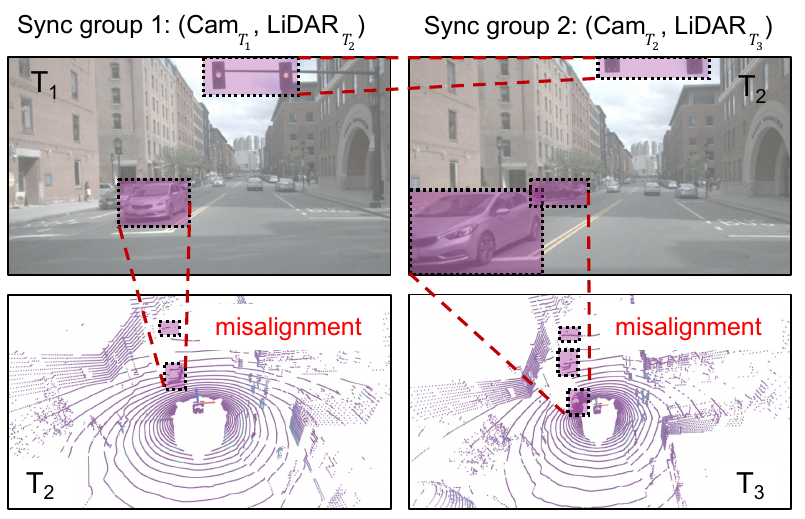}
    \vspace{-5mm}
    \caption{An example of camera–LiDAR misalignment. Two sync groups 
    are formed: 
    $\{T_1^{\text{cam}},T_2^{\text{lidar}}\}$ and 
    $\{T_2^{\text{cam}},T_3^{\text{lidar}}\}$ with
    $\Delta t_{1,2}{=}350$\,ms and $\Delta t_{2,3}{=}150$\,ms. 
    Objects move and FOV content changes across these offsets, 
    leading to conflicting cross-modal evidence.}
   \vspace{-3mm}
    \label{fig:misalignment}
\end{figure}

\vspace{1mm}
\noindent \textbf{Misalignment measurably degrades fusion.}
Figure~\ref{fig:misalignment} illustrates two synchronization groups 
produced by the default ROS \texttt{ApproximateTime}: 
(i) $\{$camera @ $T_1$, LiDAR @ $T_2\}$ and 
(ii) $\{$camera @ $T_2$, LiDAR @ $T_3\}$, 
where $\Delta t_{1,2}=350$\,ms and $\Delta t_{2,3}=150$\,ms. 
Between $T_1$ and $T_2$, the lead vehicle moves closer and a traffic light 
goes out of the camera’s FOV; fused with LiDAR @ $T_2$, the appearance/geometry 
no longer exists, creating a feature conflict. 
Between $T_2$ and $T_3$, LiDAR captures \emph{three} vehicles while the 
camera sees \emph{two}, causing ghosting/duplication or misses during fusion. 
These disagreements do not stem from model capacity, but from 
\emph{time}—--naïvely aligned features describe different physical 
states of the world.

\vspace{1mm}
\noindent \textbf{Aligning to low-rate keyframes is not enough.}
A common workaround is to downsample all inputs to low-rate keyframes. 
While this reduces the nominal offset, it sacrifices temporal resolution
and reaction time, and \emph{still} leaves residual misalignment once
multi-sweep aggregation and real bus jitter are considered. 
Moreover, coarse, frame-level synchronization ignores where the 
informative content actually lies 
(Fig.~\ref{fig:probability-heatmap-nuscenes}): most camera pixels are 
low-saliency, so forcing strict frame-level alignment wastes bandwidth 
and compute on uninformative regions.


\vspace{0.5em}
\noindent\textbf{Insight 2.} \textit{Asynchronous sampling routinely
introduces $50$--$500$\,ms offsets between camera and LiDAR. Naively
aligned features describe different physical states of the world,
producing feature conflicts (ghosting, mismatched object sets) that
degrade fusion accuracy.}

\subsection{Why Criticality?}

\begin{figure}[h]
    \centering
    \includegraphics[width=\columnwidth]{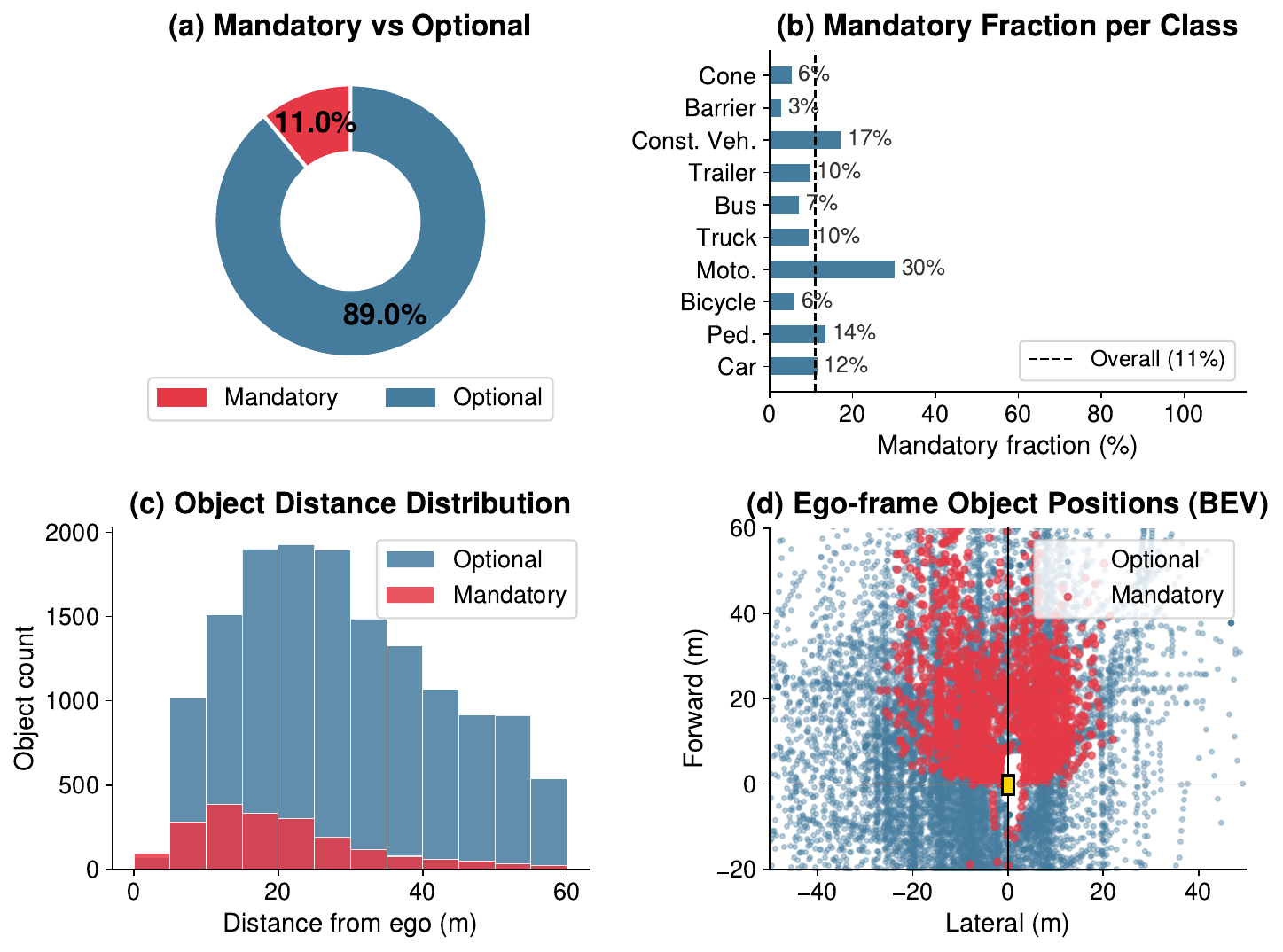}
    \vspace{-4mm}
    \caption{Criticality analysis on nuScenes v1.0-mini ($18{,}364$ annotations).
    (a) Only 11\% of objects are mandatory (safety-critical); 89\% are
    optional. (b) Per-class mandatory fractions. (c) Critical objects
    concentrate at close range; optional ones span the full detection
    envelope. (d) In BEV, mandatory objects (red) cluster in the ego's
    forward path while optional objects (blue) populate the periphery.}
    \vspace{-2mm}
    \label{fig:criticality-overview}
\end{figure}

\noindent\textbf{Most detected objects are safety-noncritical.}
Standard BEV evaluation measures recall over \emph{all} annotated
objects---pedestrians 50\,m behind the ego, traffic cones on the
opposite shoulder, parked vehicles far outside any plausible collision
path. These objects contribute equally to mAP, yet a planner would
not alter its immediate control action for any of them.
To quantify this structure, we classify each annotated object as
\emph{mandatory} or \emph{optional} based on the ego's driving state:
an object is \emph{mandatory} if it lies inside the ego's forward
braking distance or its time-to-collision falls below a per-class
threshold, and \emph{optional} otherwise (formalized in
\S\ref{sec:roi-selector}). Applying this classifier to all $18{,}364$
annotations across the nuScenes v1.0-mini scenes,
Fig.~\ref{fig:criticality-overview} shows that only $\mathbf{11.0\%}$
of all annotations are mandatory; the remaining $\mathbf{89.0\%}$ are
optional. The mandatory fraction varies by class---motorcycles
($30.4\%$) and construction vehicles ($17.3\%$) are most likely to be
in the ego's path, while barriers ($3.0\%$) and traffic cones
($5.5\%$) rarely are---but no class exceeds one third of its
annotations being mandatory.

\vspace{0.5em}
\noindent\textbf{Insight 3.} \textit{An overwhelming majority of
annotated objects ($\approx\!89\%$ on nuScenes) fall outside the ego's
braking envelope and short-TTC window, so they are irrelevant to the
planner's immediate control action. All-object recall therefore
conflates two very different costs: dropping an optional far-field
object is safety-neutral, while dropping a mandatory in-path object
is not. Real-time BEV perception must be designed and \emph{measured} against
the mandatory subset.}


\section{System Design}
\label{sec:design}

This section details the design of \name{}. Building on the three
insights of Section~\ref{sec:empirical-study}, \name{} routes both
modalities' compute through fine-grained ROIs (addressing the
spatial/temporal redundancy of Insight~1), treats synchronization and
detection as a joint problem to mitigate the inter-modal timing
offsets we measured (Insight~2), and decomposes perception into a
\emph{mandatory} part --- the safety-critical detections a planner
must act on this cycle --- and an \emph{optional} part (Insight~3),
prioritizing the mandatory part for low-latency completion and
shedding optional work when the compute budget is tight. We first give
an overview, then state the technical challenges, then detail each
component.

\subsection{System Overview}

\begin{figure}[h]
    \centering
    \includegraphics[width=\columnwidth]{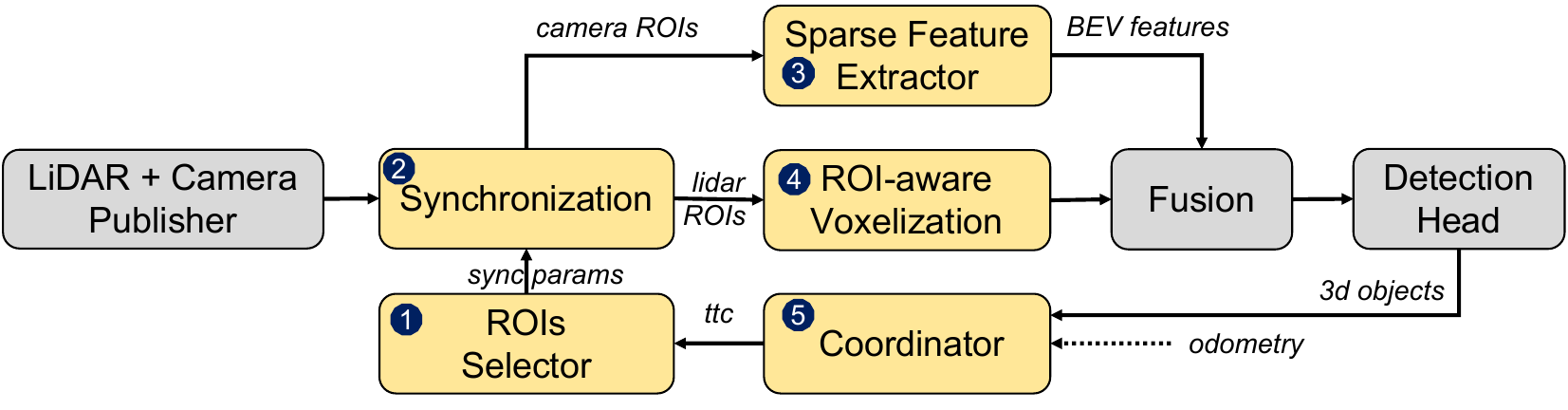}
    \vspace{-4mm}
    \caption{\name{} system overview.}
    \vspace{-2mm}
    \label{fig:system-design}
\end{figure}

Fig.~\ref{fig:system-design} depicts the architecture of \name{}. Unlike the
conventional pipeline of Fig.~\ref{fig:bev-pipeline}, which densely extracts
features from every frame, \name{} routes compute by \emph{where it matters}
and \emph{when it matters}. It begins with a \emph{criticality-ranked
temporal-ROI selector} \circled{1}: rather than re-scanning each frame, it
projects the previous frame's motion-extrapolated detections into
per-modality regions of interest (ROIs) and ranks them by criticality, so
the mandatory set is always covered while redundant coverage is dropped.
A \emph{synchronization module} \circled{2} assembles the raw sensor data
for the selected ROIs from a buffer and motion-compensates multi-sweep LiDAR
to a common reference time. The camera branch performs \emph{sparse
ROI-aware feature extraction} \circled{3}, running the backbone only on a
shared-shape ROI crop at a context-adaptive resolution; the LiDAR branch
uses \emph{ROI-aware voxelization} \circled{4}, voxelizing only the ROI
point set. At the core, a \emph{latency-aware coordinator} \circled{5} holds
the per-frame budget --- scheduling the LiDAR sweep window, the image
resolution, and periodic full-scene keyframes from scene dynamics and
time-to-collision (TTC) --- and runs the whole pipeline under an
\emph{asynchronous producer/consumer scheduler} so the model always
consumes the freshest synchronized sensor bundle.

\subsection{Technical Challenges}

\name{} is designed to address four challenges in achieving efficient
and accurate BEV perception with real-time latency.

\vspace{1mm}
\noindent\textbf{C1: Modeling where compute matters.}
\textit{How can the scene be modeled to route compute to the regions that
matter, without missing safety-critical objects?} \name{} exploits temporal
coherence --- the previous frame's detections, constrained by physical
motion, localize this frame's objects --- and ranks the resulting ROIs by
\emph{criticality} derived from ego state, so the mandatory set is never
dropped.

\vspace{1mm}
\noindent\textbf{C2: Mitigating sensor misalignment.}
\textit{How can temporal misalignment across asynchronous sensors be
minimized?} \name{} forms ROI-guided synchronization groups and
motion-compensates multi-sweep LiDAR to a common reference time before
fusion.

\vspace{1mm}
\noindent\textbf{C3: Processing sparse, variable-sized inputs.}
\textit{How can feature extraction handle sparse ROIs efficiently on
encoders built for dense inputs?} \name{} runs the camera backbone on a
shared-shape ROI crop at a context-adaptive resolution and confines LiDAR
voxelization to the ROI point set --- limiting computation to informative
regions without padding waste.

\vspace{1mm}
\noindent\textbf{C4: Keeping latency low.}
\textit{How can the latency--accuracy trade-off be managed so the pipeline
sustains real-time rates?} A latency-aware coordinator schedules
keyframes, the LiDAR sweep window, and image resolution from scene
dynamics and TTC, and an asynchronous scheduler keeps end-to-end latency
low by always running on the freshest sensor bundle.

\subsection{Criticality-Ranked Temporal-ROI Selector}
\label{sec:roi-selector}

The selector reduces redundant computation by extracting sparse, semantically
critical regions from raw sensor data. It rests on two ideas: \emph{temporal
localization} --- the previous frame tells us where objects are --- and
\emph{criticality ranking} --- which of those regions are mandatory.

\vspace{1mm}
\noindent\textbf{Temporal ROI localization.} Object motion between
consecutive frames is physically bounded, so the previous frame's detections
are a reliable prior for the current frame. Let $\mathcal{B}^{t-1}$ be the
set of boxes detected at frame $t-1$, each carrying an estimated velocity
$v_i$. \name{} projects every box forward by $v_i\,\Delta t$ and forms
per-modality ROIs from the union of current and motion-extrapolated box
footprints. For the LiDAR branch, ROIs are axis-aligned BEV bounds buffered
by a margin $r$:
\begin{align}
\mathrm{ROI_L} = \bigcup_{b_i \in \mathcal{B}^{t-1}}
  \bigl( (b_i \,\cup\, (b_i + v_i\,\Delta t)) \oplus r \bigr).
\end{align}
For the camera branch, the same box set is projected to each image plane and
expanded by $\epsilon$ pixels for localization uncertainty:
\begin{align}
\mathrm{ROI_C} = \bigcup_{b_i \in \mathcal{B}^{t-1}}
  \Bigl(\mathrm{Proj}_{2D}(b_i + v_i\,\Delta t) \oplus \epsilon\Bigr).
\end{align}
The image ROIs across all cameras are normalized to a single
\emph{shared shape} (\S\ref{sec:sparse-camera}) so the backbone forward is
batched without padding waste; cameras with no projected box receive a small
fallback crop. When the cache is empty (first frame, or after a keyframe),
the selector falls back to the full-frame pipeline.

\vspace{1mm}
\noindent\textbf{Criticality model.} Not every detected object carries
equal real-time value. \name{} scores each object from the ego state ---
speed $v_{\rm ego}$ and heading --- and the object's geometry, into two
classes. For an object with relative position
$\mathbf{p}_i = (x_i, y_i)$ ($x_i$ forward of the ego, $y_i$ lateral)
and class $c(i)$, the object is \emph{geometry-critical} if it lies
within the ego's forward braking cone:
\begin{align}
\label{eq:geom-crit}
\mathrm{Geom}(i) \;\Leftrightarrow\;
x_i \!\ge\! 0 \,\wedge\,
\|\mathbf{p}_i\| \!\le\! d_{\max}(v_{\rm ego}) \,\wedge\,
\big|\!\arctan\!\tfrac{y_i}{x_i}\big| \!\le\! \theta(v_{\rm ego}),
\end{align}
with braking distance
\begin{align}
\label{eq:dmax}
d_{\max}(v_{\rm ego}) = \max\!\left(d_{\min},\; \frac{v_{\rm ego}^2}{2\,a_{\rm brake}}\right),
\end{align}
where $\theta(v_{\rm ego})$ is the forward-cone half-angle at the
current ego speed, $a_{\rm brake}$ is a nominal deceleration, and
$d_{\min}$ a floor for low-speed proximity. The object is
\emph{safety-critical} if its time-to-collision
(\S\ref{sec:coordinator}, Eq.~\ref{eq:ttc}) falls below a per-class
threshold $\tau_{c(i)}$:
\begin{align}
\label{eq:safety-crit}
\mathrm{Safety}(i) \;\Leftrightarrow\; \mathrm{TTC}_i \le \tau_{c(i)},
\end{align}
with $\tau_c$ tighter for vulnerable road users than for static
objects. The \emph{mandatory} set is the union
$\mathcal{M} = \{i : \mathrm{Geom}(i) \vee \mathrm{Safety}(i)\}$; the
rest is \emph{optional}. Criticality ranks the ROIs: the selector
guarantees coverage of mandatory ROIs at full fidelity and is free to
shrink, down-sample, or drop optional ROIs to hold the latency
budget (\S\ref{sec:coordinator}). The criticality model is also the
basis of the evaluation metric (\S\ref{sec:eval}), which measures
accuracy on the mandatory set $\mathcal{M}$ rather than over all
objects.

\vspace{1mm}
\noindent\textbf{Criticality thresholds.} The forward-cone half-angle
$\theta(v_{\rm ego})$ is a piecewise-linear function of ego speed: it
widens at low speed (where lateral maneuvers are still possible) and
narrows at highway speed (where the relevant corridor is forward and
narrow), bounded between $15^\circ$ and $60^\circ$. Per-class TTC
thresholds $\tau_c$ are set conservatively: $1.5$\,s for vulnerable road
users (pedestrians, cyclists) and $2.5$\,s for motorized objects; static
infrastructure (barriers, cones) is treated as geometry-critical only,
since their relative velocity is zero and TTC is undefined. These values
are tunable deployment parameters matched to typical highway-code
reaction-distance requirements.



\subsection{Synchronization}

Synchronization integrates sensor data across time and space. \name{} treats
synchronization and fusion as one module, aligning data temporally and
spatially before fusion.

\vspace{1mm}
\noindent\textbf{ROI-guided sync groups.} Not all sensor data is equally
informative at every timestep~\cite{yan2023cross}. \name{} adapts the
synchronization group from ROI information rather than always fusing all six
cameras and LiDAR~\cite{liu2024rt}, and occasionally promotes a
comprehensive keyframe involving all sensors (\S\ref{sec:coordinator}). The
synchronization window is set from the ego speed and TTC.

\begin{figure}[h]
\centering
\includegraphics[width=\columnwidth]{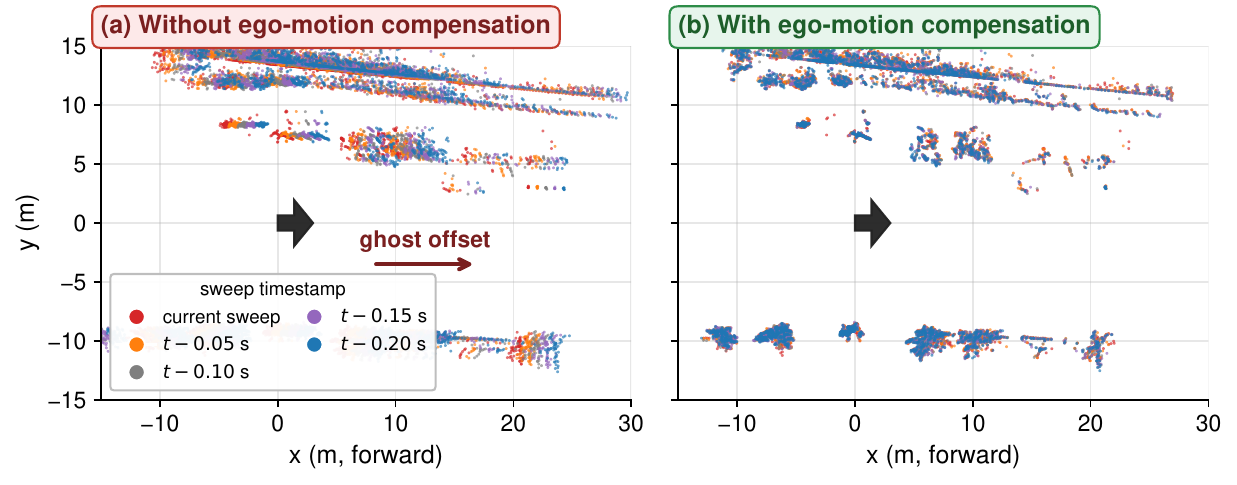}
\vspace{-4mm}
\caption{Aggregation of LiDAR sweeps from a nuScenes scene without
(a) and with (b) ego-motion compensation, above-ground returns only.
Color encodes sweep age (red = current sweep, blue = oldest).}
\label{fig:motion-compensation}
\vspace{-2mm}
\end{figure}

\vspace{1mm}
\noindent\textbf{Motion compensation.} Aggregating multiple LiDAR
sweeps introduces time differences that misalign
points~\cite{xia2025robust}, as shown in
Figs.~\ref{fig:lidar-sweeps-map-time} and \ref{fig:misalignment}.
\name{} compensates with the ego motion between sweep timestamps: a
point from an earlier sweep at $t-\Delta t$ is transformed to the
current frame as
\begin{align}
\mathbf{p}_{t-\Delta t}^{\prime} = \mathbf{R}_{t-\Delta t}^{t}
\mathbf{p}_{t-\Delta t} + \mathbf{T}_{t-\Delta t}^{t},
\end{align}
and the current sweep is aggregated with $n$ motion-compensated
previous sweeps,
\begin{align}
\mathcal{P}_t = \bigcup_{i=0}^{n} \left\{ \mathbf{R}_{t-i}^{t}
\mathbf{p}_{t-i} + \mathbf{T}_{t-i}^{t} \,\middle|\,
\mathbf{p}_{t-i} \in \mathcal{P}_{t-i} \right\},
\end{align}
where $n$ is the sweep count chosen by the coordinator
(\S\ref{sec:coordinator}). Fig.~\ref{fig:motion-compensation}
illustrates the effect on a nuScenes scene with ego motion across the
sweep window: without compensation, static structures appear as
several ghost copies offset along the direction of travel; with
compensation, the same structures collapse to a single coherent
position.

\subsection{Sparse ROI-Aware Camera Feature Extraction}
\label{sec:sparse-camera}

Image ROIs occupy only a small fraction of the dense image, so the
camera backbone should not process every pixel.

\vspace{1mm}
\noindent\textbf{Shared-shape ROI crop.} \name{} crops each camera to its
ROI window and normalizes all crops to a single \emph{shared shape}
$(\textit{target}_h, \textit{target}_w)$, sized from the largest projected
ROI and snapped to the backbone stride. Per-camera offsets place each crop;
cameras with no ROI get a small center crop. Because all crops share a
shape, the backbone forward is a
single batched pass with no padding waste. After the backbone and neck,
the cropped features are pasted onto a full-image feature canvas at the
correct stride, and a binary \emph{ROI mask} marks the pasted region.
The mask is carried into the view transform and zeroes the BEV
contribution of canvas regions that held no real features, removing the
camera-side accuracy floor that a zero-padded canvas would otherwise
introduce.

\vspace{1mm}
\noindent\textbf{Context-adaptive resolution.} The camera backbone is the
dominant cost in the image branch. \name{} runs it at a context-adaptive
resolution: when the scene is slow
and sparse, the ROI crop is downsampled before the backbone and the
features are upsampled back, halving the image backbone cost; when the
scene is fast and dense, full resolution is kept. The resolution choice
is made per frame by the coordinator (\S\ref{sec:coordinator}).


\subsection{ROI-Aware Voxelization}

LiDAR point clouds are sparse, yet conventional pipelines voxelize the whole
space and run sparse convolutions over every non-empty voxel, including
regions of little semantic value~\cite{liu2023bevfusion}. \name{} confines
LiDAR processing to the ROIs.

\begin{figure}[h]
\centering
\includegraphics[width=.8\columnwidth]{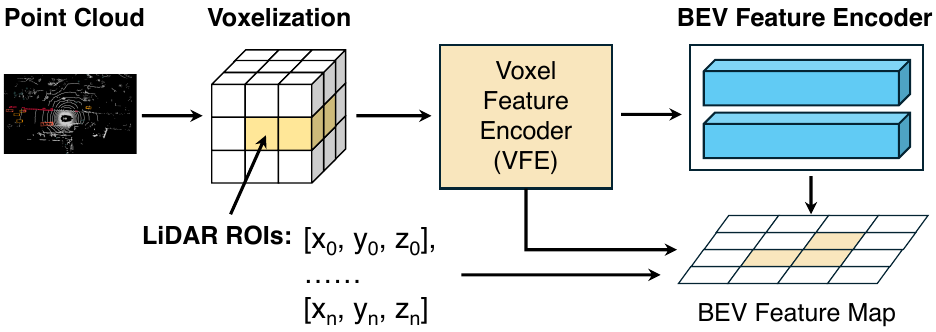}
\vspace{-4mm}
\caption{ROI-aware voxelization and feature extraction.}
\vspace{-2mm}
\label{fig:ROIs-PCD}
\end{figure}

As in Fig.~\ref{fig:ROIs-PCD} and Alg.~\ref{alg:lidar-downsample}, \name{}
keeps points by their position relative to $\mathrm{ROI_L}$: points inside
the (motion-padded) ROI box union are kept in full, and points outside are
kept at a reduced rate, so most of the background is dropped while the ROI
neighborhood --- including localization slack --- is preserved. The retained
points are voxelized and passed through the Voxel Feature Encoder; features
are scattered into a zero-initialized BEV map by voxel coordinate. The
result has the same BEV shape as a dense pipeline but is computed from a
focused subset of points, cutting LiDAR cost without sacrificing coverage of
critical regions.

\begin{algorithm}[h]
\caption{ROI-Aware Voxelization}
\label{alg:lidar-downsample}
\small
\begin{algorithmic}[1]
\Function{ROIAwareVoxelization}{$points$, $\mathrm{ROI}_L$}
\State $kept \gets$ []
\For{each $point$ in $points$}
\If{$point$ inside any box in $\mathrm{ROI}_L$}
  \State keep $point$ \Comment{inside-ROI rate}
\Else
  \State keep $point$ with probability $\rho_{\rm out}$
\EndIf
\EndFor
\State $voxels, coords \gets$ \textbf{Voxelize}($kept$)
\State $features \gets$ \textbf{VoxelFeatureEncoder}($voxels$)
\State \Return \textbf{ScatterToBEV}($features$, $coords$)
\EndFunction
\end{algorithmic}
\end{algorithm}

\begin{figure}[h]
\centering
\includegraphics[width=\columnwidth]{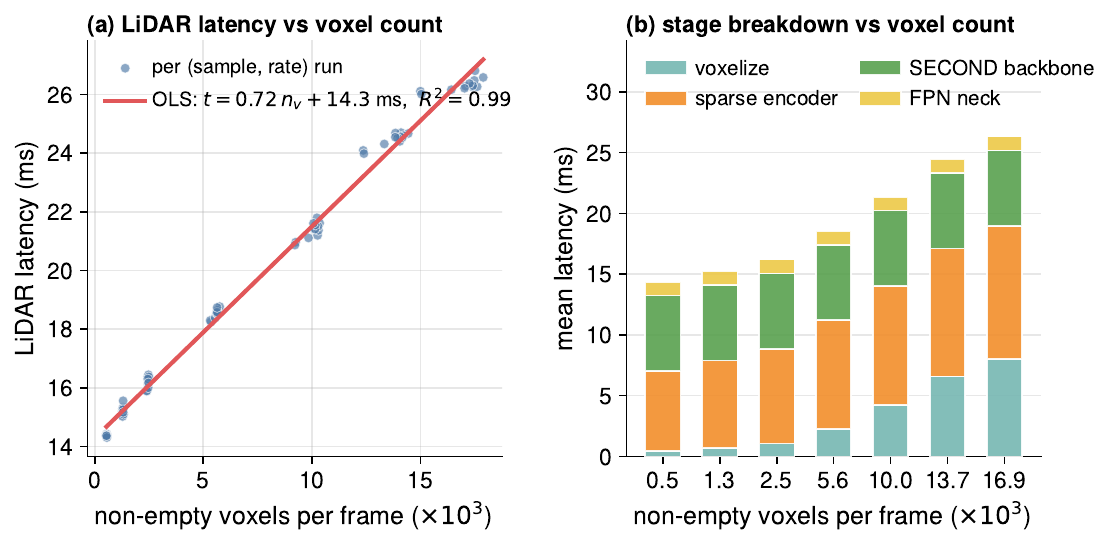}
\vspace{-5mm}
\caption{LiDAR latency vs.\ non-empty voxel count, profiled on the
BEVFusion LiDAR-only branch over $84$ runs ($12$ nuScenes scenes
$\times\,7$ sub-sample rates).
(a) Total LiDAR latency is linear in $n_v$ ($R^2{=}0.99$).
(b) Voxelize and the sparse encoder scale with $n_v$.}
\label{fig:voxel-timing}
\vspace{-2mm}
\end{figure}

\noindent\textbf{LiDAR latency scales with voxel count.}
We confirm that reducing the voxel count translates directly into
inference-time savings.
We profile the BEVFusion LiDAR-only branch on $12$ nuScenes scenes,
sub-sampling each point cloud at rates
$\{1.0, 0.75, 0.5, 0.25, 0.10, 0.05, 0.02\}$ to generate a range of
voxel counts after hard voxelization, and time each LiDAR stage on the
RTX\,3080.
Across $84$ runs (Fig.~\ref{fig:voxel-timing}), the total LiDAR
latency is well-modeled by a linear fit
$t \approx 0.72\,n_v + 14.3$\,ms ($R^2{=}0.99$), where $n_v$ is the
non-empty voxel count in thousands.
The breakdown shows that voxelization and the sparse encoder grow
nearly linearly with $n_v$, while the SECOND backbone and FPN neck
stay constant at $\sim\!7$\,ms.
A full-resolution sweep yields $\sim\!17$\,k voxels and $26$\,ms;
restricting compute to the ROI brings $n_v$ to $\sim\!5$\,k and the
LiDAR cost to $18$\,ms---a $1.4\times$ saving on this stage alone.

\vspace{-2mm}
\subsection{Latency-Aware Coordinator}
\label{sec:coordinator}

The coordinator holds the per-frame latency budget. It does so along two
axes: \emph{compute scaling} --- adjusting how much work each frame does ---
and \emph{pipeline scheduling} --- ensuring the model always runs on fresh
data.

\vspace{1mm}
\noindent\textbf{Time-to-collision.} The coordinator's signals come from
ego/object motion. Given an obstacle's relative position
$\mathbf{p}=[x_1,y_1]$ and velocity $\mathbf{v}=[v_x-v_{\rm ego},v_y]$, the
time of closest approach is
\begin{align}
\label{eq:ttc}
t^{*} = -\frac{\mathbf{p}\cdot\mathbf{v}}{\lVert\mathbf{v}\rVert^{2}},
\end{align}
and $\mathrm{TTC}=t^{*}$ if $t^{*}>0$ and
$\lVert\mathbf{p}+t^{*}\mathbf{v}\rVert \le R_{\rm coll}$, else
$\mathrm{TTC}=\infty$. The minimum TTC over the scene drives both the
safety-criticality test (\S\ref{sec:roi-selector}) and the coordinator's
budget decisions.

\vspace{1mm}
\noindent\textbf{Priority-based budget management.} The coordinator
has two compute knobs---the LiDAR \emph{sweep count} $n$ and the image
\emph{resolution}---and applies them in priority order. Mandatory-ROI
coverage is fixed first and cannot be shed. The sweep count is reduced
next, from $n{=}10$ at high ego speed down to $n{=}5$ when the ego is
slow or the scene is dense (recent frames already carry the needed
context). Image resolution is dropped last, to half when the scene is
slow and sparse and kept full otherwise. If the minimum configuration
still risks running long, the coordinator falls back to LiDAR-only
inference for that frame, preserving the mandatory LiDAR-ROI path
while shedding the camera branch entirely. Both knobs trade optional
fidelity for latency and never touch the mandatory ROIs; on the
evaluation platform, the half-resolution $n{=}5$ ROI path consistently
fits within the ${\sim}100$\,ms per-frame target.

\vspace{1mm}
\noindent\textbf{Keyframes.} ROI-based perception risks error accumulation
and stale context, since newly appearing objects have no temporal
precedent. \name{} periodically inserts a \emph{keyframe} --- a full-scene
pass that refreshes the detection cache and bounds the worst-case staleness
of the temporal prior. The keyframe interval may be fixed or triggered by
scene dynamics: \name{} estimates the per-frame ROI change
\begin{align}
\Delta\mathrm{ROI}_L &= \tfrac{1}{N_L}\!\sum_{i=1}^{N_L}
  |A_{L,i}^{t}-A_{L,i}^{t-1}|, \\
\Delta\mathrm{ROI}_C &= \tfrac{1}{N_C}\!\sum_{i=1}^{N_C}
  |(c_{C,i}^{t}-c_{C,i}^{t-1}) - v_{\rm ego}\,\Delta t|,
\end{align}
where $A$ is a LiDAR-ROI area and $c$ an ego-motion-compensated camera-ROI
centroid; a keyframe is forced when either exceeds a profiled threshold.

\vspace{1mm}
\noindent\textbf{Asynchronous scheduling.} Compute scaling reduces per-frame
work, but in a live pipeline it does not by itself bound end-to-end latency:
if inference is invoked synchronously inside the sensor callback, frames
queue and the detector consumes ever-staler data. \name{} runs the pipeline
as a \emph{producer/consumer} pair: the sensor callback only buffers the
latest synchronized bundle and returns immediately, while a separate
inference thread always consumes the freshest buffered bundle and discards
stale ones. With a single-slot buffer, an unbounded backlog becomes a
bounded queue and end-to-end latency becomes predictable. Algorithm
~\ref{alg:coordinator-module} summarizes the per-frame coordination.

\begin{algorithm}[h]
\caption{Coordination}
\label{alg:coordinator-module}
\small
\begin{algorithmic}[1]
\Function{Coordinate}{$bundle$, $\mathcal{B}^{t-1}$, $v_{\rm ego}$}
  \State $bundle \gets$ \textbf{LatestBuffered}() \Comment{freshest-wins}
  \If{keyframe due \textbf{or} $\Delta\mathrm{ROI} > \tau$}
    \State \Return \textbf{FullScenePass}($bundle$)
  \EndIf
  \State $\mathrm{ROI_L},\mathrm{ROI_C} \gets$ \textbf{TemporalROI}($\mathcal{B}^{t-1}$)
  \State $\mathrm{TTC} \gets$ \textbf{ComputeTTC}($\mathcal{B}^{t-1}$, $v_{\rm ego}$)
  \State $n,\,res \gets$ \textbf{ScaleCompute}($v_{\rm ego}$, density, $\mathrm{TTC}$)
  \State $BEV_C \gets$ \textbf{SparseCamera}($\mathrm{ROI_C}$, $res$)
  \State $BEV_L \gets$ \textbf{ROIVoxelize}($\mathrm{ROI_L}$, $n$ sweeps)
  \State \Return \textbf{Detect}(\textbf{Fuse}($BEV_C$, $BEV_L$))
\EndFunction
\end{algorithmic}
\end{algorithm}

\section{Implementation}
\label{sec:implementation}

We integrate \name's implementation into BEVFusion's perception 
pipeline using ROS on a resource-limited GPU desktop
\cite{liu2023bevfusion, quigley2009ros}. 
BEVFusion is a widely deployed baseline for multi-modality BEV 
perception~\cite{liu2023bevfusion, liu2021swin, xie2023sparsefusion, bai2022transfusion, liang2022bevfusion}.

\begin{figure}[h]
\centering
\includegraphics[width=\columnwidth]{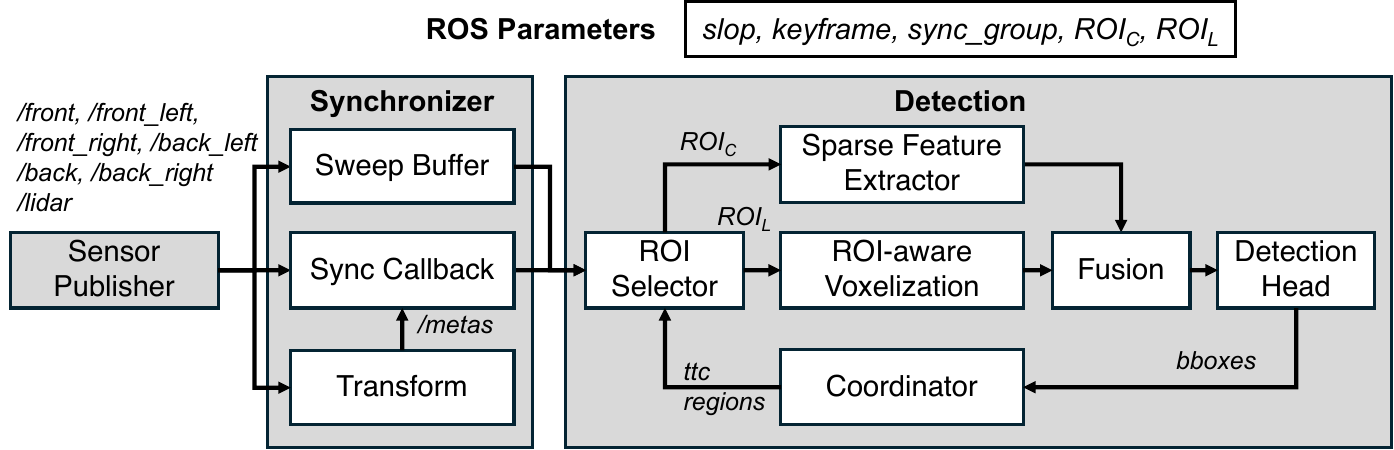}
\vspace{-5mm}
\caption{ROS framework for \name.}
\vspace{-2mm}
\label{fig:ROS-framework}
\end{figure}

\vspace{1mm}
\noindent\textbf{ROS framework.} To emulate real-time streaming 
processing similarly to actual AVs, we 
implement \name{} using ROS~\cite{quigley2009ros, message-filter}, 
as depicted in Fig.~\ref{fig:ROS-framework}. 
Specifically, we develop three ROS nodes: \texttt{Sensor Publisher}, 
\texttt{Synchronizer}, and \texttt{Detector}. 
The \texttt{Sensor Publisher} publishes ROS bags generated
from nuScenes dataset,
incorporating raw sensor data (6 cameras and 1 LiDAR) 
alongside CAN bus information from the ego vehicle. Camera frames are 
published at 12 Hz, while LiDAR data is published at 20 Hz. 
The \texttt{Synchronizer} node consists of a \texttt{Sweep Buffer} 
that caches previous LiDAR sweeps, a \texttt{Transform} module for 
precomputing metadata containing sensor coordinate transformations 
(rotation and translation), and a \texttt{Sync Callback}, which serves as 
the entry point for synchronizing sensor data based on the ROIs within a 
configurable \texttt{slop} threshold. The \texttt{Detector} node closely
follows the design structure of \name, with minor adaptations for ROS-based
inter-module communication using ROS parameters. Specifically, parameters
like \texttt{slop} and \texttt{sync\_group} are employed by the
\texttt{Coordinator} to dynamically adjust synchronization settings.
The \texttt{keyframe} parameter also indicates the keyframe mode,
and \texttt{$ROI_C$} and \texttt{$ROI_L$} parameters are used for
real-time updates of ROIs for cameras and LiDAR, respectively.

\vspace{1mm}
\noindent\textbf{Asynchronous scheduling.}
The \texttt{Synchronizer} publishes each synchronized bundle to a
single-slot topic and returns immediately; the \texttt{Detector}
subscribes with \texttt{queue\_size=1} and an overwrite-on-arrival
callback, so a bundle arriving while inference is busy replaces the
queued one rather than being appended. The \texttt{Detector} thus
always consumes the freshest available bundle.

\section{Evaluation}
\label{sec:eval}

We evaluate \name{} end-to-end on the nuScenes dataset replayed
through ROS and via a controlled offline timing benchmark,
then deploy the system on a real robot.

\subsection{Experimental Setup}
\label{sec:setup}

\vspace{1mm}
\noindent\textbf{Hardware.}
Offline timing benchmarks run on a desktop with a 12th-Gen
Intel\textregistered{} Core\texttrademark{} i9-12900K (3.3\,GHz) and
an NVIDIA GeForce RTX\,3080 (29.8\,TFLOPS FP32, 10\,GB GDDR6).
End-to-end ROS experiments run on the same machine.
Libraries: \texttt{CUDA}\,12.3, \texttt{PyTorch}\,2.4.1,
\texttt{mmdet3d}\,1.2.0, \texttt{ROS\,Noetic}.

\vspace{1mm}
\noindent\textbf{Dataset.}
We use the nuScenes dataset~\cite{caesar2020nuscenes}:
11,628 images (1600$\times$928, 6 cameras) plus 20-sweep LiDAR,
covering 10 driving scenes.
In the ROS experiments, camera frames are published as
\texttt{sensor\_msgs/Image} at 12\,Hz and LiDAR as
\texttt{sensor\_msgs/PointCloud2} at 20\,Hz.

\vspace{1mm}
\noindent\textbf{Baselines and configurations.}
Table~\ref{tab:configs} lists all evaluated configurations.
\texttt{BEVFusion}~\cite{liu2023bevfusion} is the primary baseline
(dense, full-resolution, synchronous pipeline).
\texttt{Scheduler\,Only} adds the asynchronous streaming scheduler
alone, without ROI or adaptive resolution.
Ablation variants progressively add ROI routing, adaptive sweep
count, and adaptive (or fixed low) image resolution.
\texttt{MM-BEV} is the full proposed system.

\begin{table}[t]
\caption{Evaluated configurations.}
\label{tab:configs}
\centering
\resizebox{\columnwidth}{!}{%
\begin{tabular}{@{}lcccc@{}}
\toprule
\textbf{Config} & \textbf{Sensors} & \textbf{Temporal ROI} &
    \textbf{Adaptive res.} & \textbf{Async sched.} \\
\midrule
BEVFusion~\cite{liu2023bevfusion} & L+C & \texttimes & \texttimes & \texttimes \\
Scheduler Only & L+C & \texttimes & \texttimes & \checkmark \\
ROI & L+C & \checkmark & \texttimes & \texttimes \\
ROI+LowRes & L+C & \checkmark & fixed-low & \texttimes \\
ROI+Adapt+LowRes & L+C & \checkmark & adaptive & \texttimes \\
\midrule
\textbf{MM-BEV} & \textbf{L+C} & \checkmark & adaptive & \checkmark \\
MM-BEV+LowRes & L+C & \checkmark & fixed-low & \checkmark \\
\bottomrule
\end{tabular}}
\end{table}

\vspace{1mm}
\noindent\textbf{Metrics.}
We report \emph{inference latency} (mean over non-warmup frames,
split by keyframe vs.\ ROI-frame mode),
\emph{end-to-end latency} (sensor exposure to detection output,
measured in ROS),
\emph{processed frames per scene}, and
\emph{criticality-bucketed recall}:
all-object recall,
geometry-critical recall (objects inside the ego braking cone),
and safety-critical recall.

\subsection{Inference Latency}
\label{sec:latency}

\begin{figure}[t]
\centering
\includegraphics[width=\columnwidth]{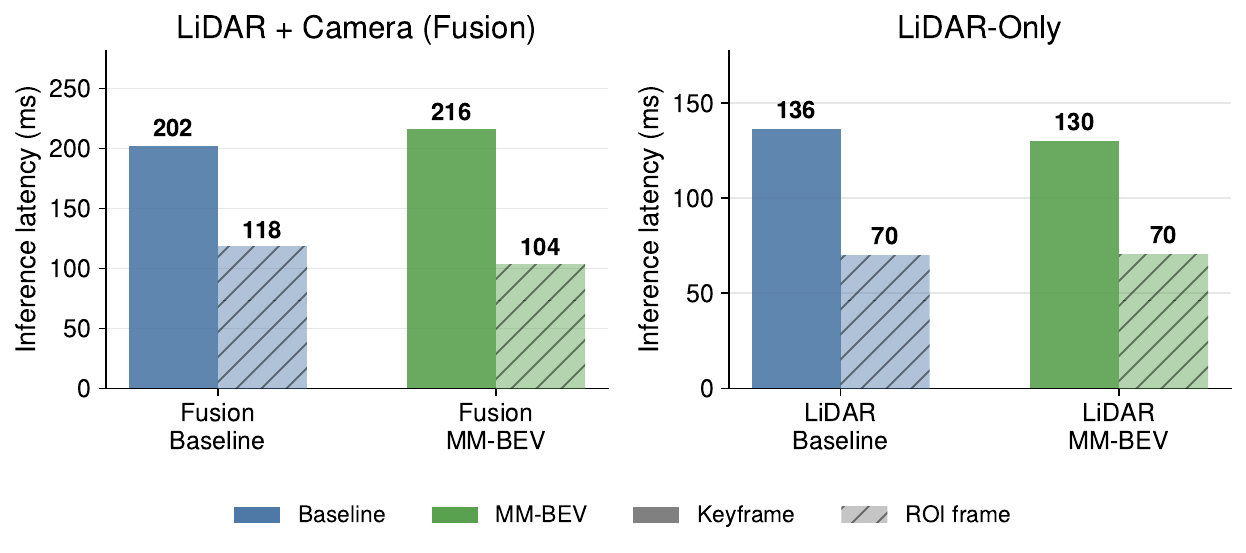}
\vspace{-5mm}
\caption{Per-frame inference latency split by keyframe and ROI-frame
mode, for the LiDAR+Camera (fusion) and LiDAR-only detectors.
Baseline must run full inference every cycle; \name{} runs fast ROI
inference on most frames and full inference only on keyframes.}
\label{fig:kf-roi-latency}
\vspace{-2mm}
\end{figure}

\begin{figure}[t]
\centering
\includegraphics[width=\columnwidth]{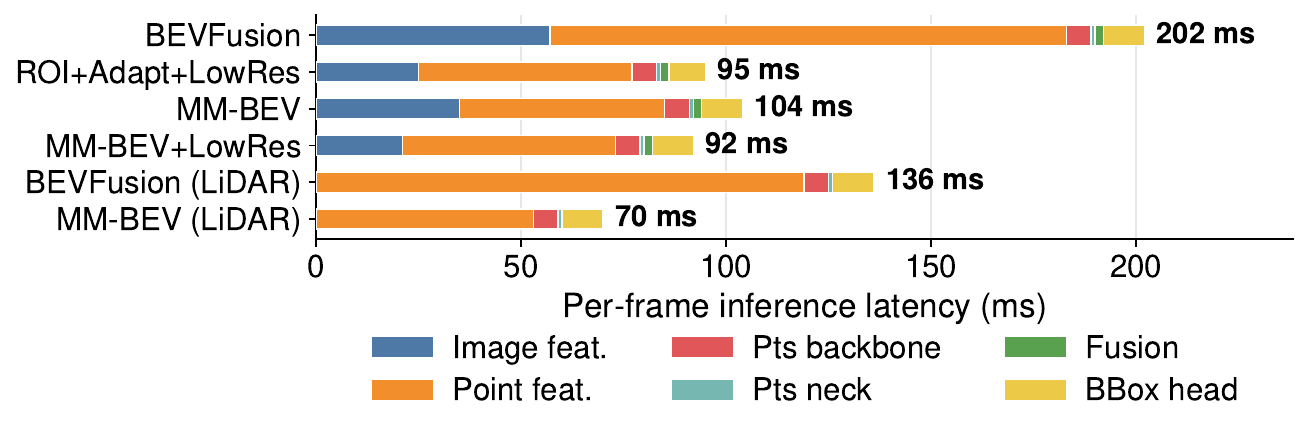}
\vspace{-5mm}
\caption{Per-stage breakdown of inference time.}
\label{fig:latency-breakdown}
\vspace{-2mm}
\end{figure}

Fig.~\ref{fig:kf-roi-latency} compares per-frame inference latency
for the fusion and LiDAR-only detectors.
\name{} classifies every frame as either a \emph{keyframe}
(full inference, triggered periodically or by high scene dynamics)
or a \emph{ROI frame} (inference restricted to the temporal ROI).

\vspace{1mm}
\noindent\textbf{Fusion detector.}
The baseline must run full-resolution, full-sweep inference on every
frame: \textbf{202\,ms} per keyframe and \textbf{118\,ms} per
``ROI frame'' (the baseline uses adaptive sweeps but no ROI crop).
\name{} cuts the ROI-frame cost to \textbf{104\,ms} (a
\textbf{1.96$\times$} speedup over the baseline keyframe), while
keyframe cost stays at 216\,ms due to the shared-shape crop overhead.
With a keyframe interval of $N{=}5$ (the default), the ROI frames
are the common case and dominate the average.

\vspace{1mm}
\noindent\textbf{LiDAR-only detector.}
The baseline keyframe costs \textbf{136\,ms}; \name{} ROI frames
cost \textbf{70\,ms} (\textbf{1.94$\times$}).
The LiDAR path benefits less from ROI-based voxelization
because point clouds are already sparse;
the main gain comes from the adaptive sweep window.

Fig.~\ref{fig:latency-breakdown} shows the stage-level
breakdown for the fusion detector.
Image feature extraction is the dominant lever:
the full-resolution backbone takes 57\,ms; temporal ROI
with low-resolution input cuts it to 21\,ms (2.7$\times$).
LiDAR feature extraction (84\,ms baseline) contracts only modestly
under ROI voxelization (83--91\,ms), consistent with its
already-sparse structure.
The BBox head and fusion stage together account for only 12\,ms
and are invariant across configurations.

\subsection{Criticality-Bucketed Accuracy}
\label{sec:crit-accuracy}

\begin{figure}[t]
\centering
\includegraphics[width=\columnwidth]{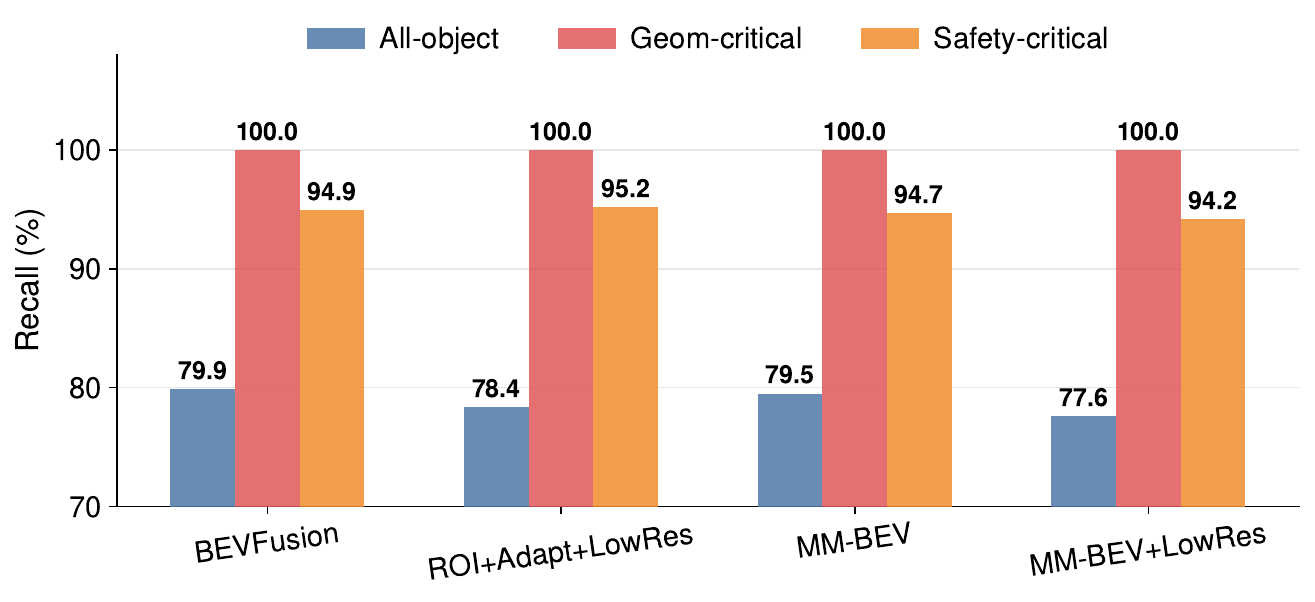}
\vspace{-5mm}
\caption{Recall by criticality bucket. Geometry-critical recall is
preserved exactly at $100\%$ across all configurations;
safety-critical recall drops at most $0.7$\,pp and all-object recall
at most $2.3$\,pp, under the most aggressive compression.}
\label{fig:crit-recall}
\vspace{-2mm}
\end{figure}

Fig.~\ref{fig:crit-recall} shows all-object, geometry-critical, and
safety-critical recall for the key configurations.

\vspace{1mm}
\noindent\textbf{Geometry-critical recall is preserved exactly.}
Every ROI-based configuration matches the baseline at
\textbf{$100\%$} on geometry-critical objects. This is by
construction: the criticality-ranked selector
(\S\ref{sec:roi-selector}) always covers the mandatory set derived
from the ego's braking cone, so no object inside that zone is skipped.

\vspace{1mm}
\noindent\textbf{Safety-critical recall degrades minimally.}
Safety-critical recall (objects below the per-class TTC threshold)
drops \textbf{$0.2$\,pp} for \name{} ($94.7\%$ vs.\ baseline $94.9\%$)
and at most \textbf{$0.7$\,pp} for \name{}+LowRes ($94.2\%$). The
small loss reflects the part of the safety-critical set that lies just
outside the braking cone and is therefore not covered by the
geometry ROI.

\vspace{1mm}
\noindent\textbf{All-object recall drops only on optional objects.}
All-object recall falls $0.4$\,pp for \name{} and $2.3$\,pp for
\name{}+LowRes; in both cases the lost detections are entirely in
the optional set (distant, low-TTC background traffic), so the
conventional all-object mAP exaggerates the safety impact of these
ROI/resolution trade-offs.

\begin{table}[t]
\caption{Accuracy vs.\ per-frame inference latency.}
\label{tab:accuracy-summary}
\centering
\resizebox{\columnwidth}{!}{%
\begin{tabular}{@{}lcccc@{}}
\toprule
\textbf{Config} & \textbf{Per-frame lat.\ (ms)} &
    \textbf{All-obj R (\%)} &
    \textbf{Geom-crit R (\%)} &
    \textbf{Safety-crit R (\%)} \\
\midrule
BEVFusion           & 202 & 79.9 & 100.0 & 94.9 \\
ROI+Adapt+LowRes    &  95 & 78.4 & 100.0 & 95.2 \\
MM-BEV              & 104 & 79.5 & 100.0 & 94.7 \\
MM-BEV+LowRes       &  92 & 77.6 & 100.0 & 94.2 \\
\midrule
LiDAR Baseline      & 136 & 76.8 & 100.0 & 93.5 \\
LiDAR MM-BEV        &  70 & 76.8 & 100.0 & 93.1 \\
\bottomrule
\end{tabular}}
\vspace{-3mm}
\end{table}

Table~\ref{tab:accuracy-summary} summarizes the latency--accuracy
trade-off. \name{}+LowRes achieves the best per-frame latency
($92$\,ms, $2.20\times$ faster than the $202$\,ms full-inference
baseline) while holding geometry-critical recall at $100\%$ and
dropping safety-critical recall by only $0.7$ points.

\subsection{Ablation Study}
\label{sec:ablation}

\begin{figure}[t]
\centering
\includegraphics[width=\columnwidth]{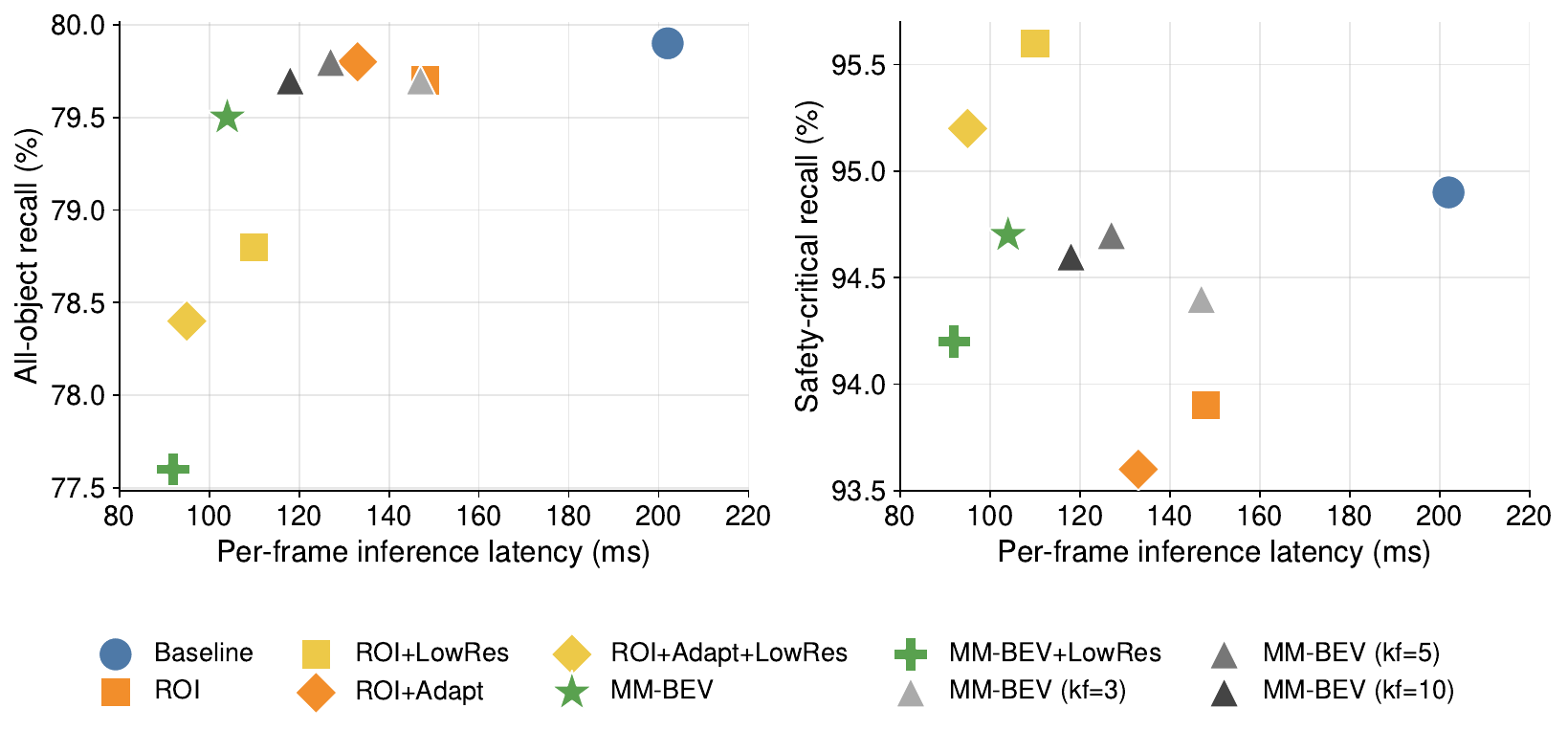}
\vspace{-5mm}
\caption{Per-frame latency vs.\ recall for all ablation configurations
(geom-critical preserved at $100\%$; right panel shows safety-critical).}
\label{fig:tradeoff}
\vspace{-2mm}
\end{figure}


\noindent\textbf{Latency--accuracy trade-off.}
Fig.~\ref{fig:tradeoff} places all ablation configurations in the
latency--recall plane, with each system at its per-frame cost in the
common-case operating mode (full inference for the baseline;
ROI-frame inference for ROI/MM-BEV variants).
Three conclusions follow.
\emph{(i) ROI alone provides limited camera savings} (ROI: $148$\,ms
ROI-frame vs.\ baseline: $202$\,ms full-inference)---the ROI crop
reshapes computation but does not reduce image resolution, so the
backbone still pays for full-resolution features.
\emph{(ii) Low resolution is the primary camera lever}
(ROI+LowRes: $110$\,ms; \name{}+LowRes: $92$\,ms ROI-frame): cutting
image input from $1600{\times}928$ to $800{\times}464$ slashes image
feature extraction from $57$\,ms to $21$\,ms.
\emph{(iii) Adaptive resolution} (\name{}: $104$\,ms ROI-frame)
keeps full resolution when scene dynamics demand it but drops to
half-res on slow/sparse scenes, maintaining all-object recall
$1.9$ points above the always-low-res variant at the cost of
$12$\,ms higher ROI-frame latency.


\begin{figure}[t]
\centering
\includegraphics[width=\columnwidth]{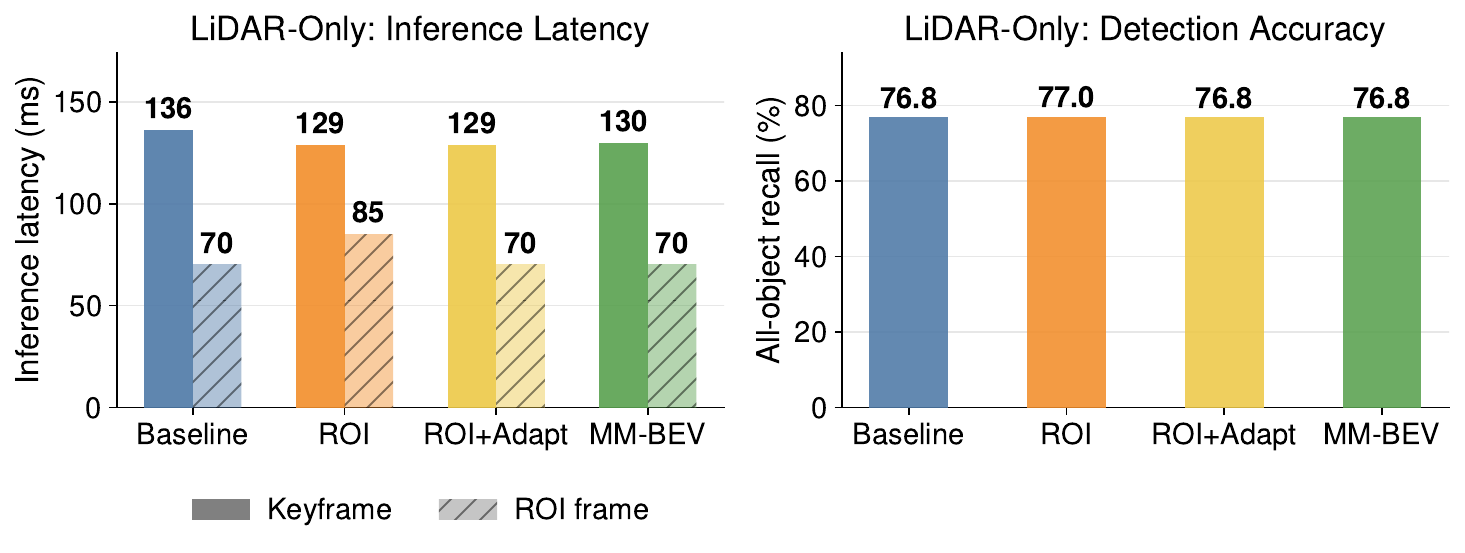}
\vspace{-5mm}
\caption{LiDAR-only detector ablation.
ROI-based voxelization alone slightly increases keyframe cost;
adaptive sweep selection reduces it.
Accuracy is stable throughout.}
\label{fig:lidar-ablation}
\vspace{-2mm}
\end{figure}

\vspace{1mm}
\noindent\textbf{LiDAR ablation.}
Fig.~\ref{fig:lidar-ablation} sweeps LiDAR-only variants.
ROI-based voxelization alone marginally increases the keyframe cost
(from 136 to 129\,ms for LiDAR-ROI), since the region selection
adds a preprocessing step without proportionally reducing the point
set on keyframes.
Adaptive sweep count (\texttt{ROI+Adapt} and \texttt{MM-BEV})
recovers this overhead, settling at 100\,ms mean---a modest
1.03$\times$ improvement.
All-object and safety-critical recall are stable across all LiDAR
configurations (76.8\% and $\sim$93\%, respectively), confirming
that the sweep-count reduction does not degrade detection quality.

\subsection{End-to-End Pipeline Performance}
\label{sec:e2e}

\begin{figure}[t]
\centering
\includegraphics[width=\columnwidth]{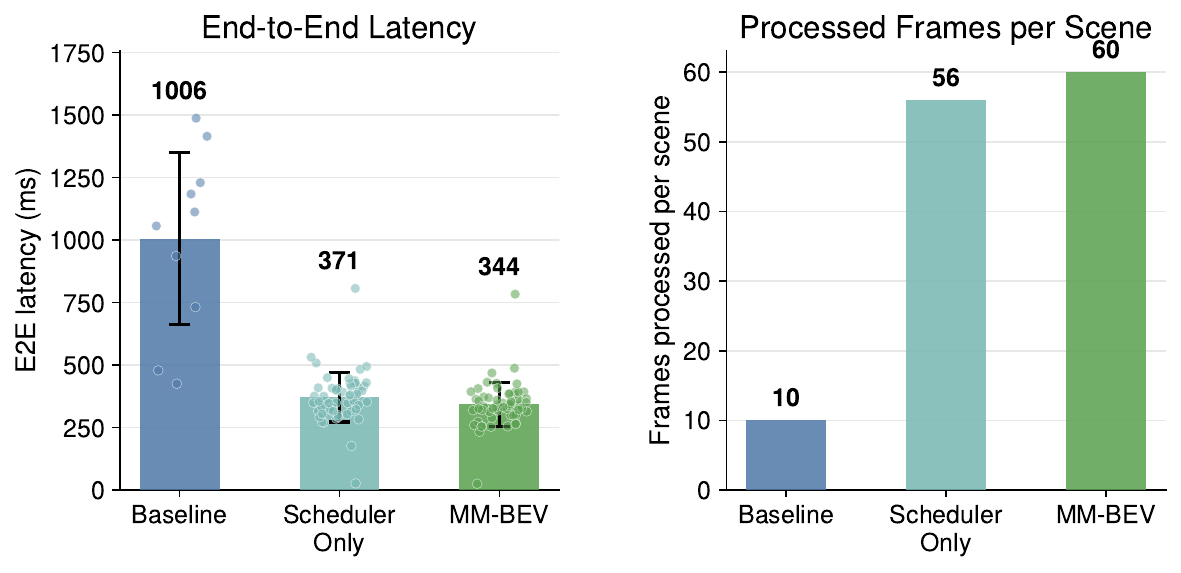}
\vspace{-5mm}
\caption{End-to-end ROS pipeline results.
\emph{Left}: mean e2e latency per processed frame (bars show mean,
error bars show $\pm$1\,s.d., jitter shows all individual frames).
\emph{Right}: total frames processed per scene.
The asynchronous scheduler is the dominant contributor to both gains.}
\label{fig:ros-e2e}
\vspace{-2mm}
\end{figure}

Fig.~\ref{fig:ros-e2e} evaluates the full pipeline in ROS, including
communication, synchronization, inference, and queuing.

\vspace{1mm}
\noindent\textbf{Scheduler dominates the e2e gain.}
The baseline pipeline stalls on its synchronous queue: camera and
LiDAR messages pile up while the detector runs at 1006\,ms e2e
latency, processing only \textbf{10\,frames} per scene.
\texttt{Scheduler\,Only}---which adds the asynchronous
producer/consumer pipeline but no ROI---already cuts e2e latency to
\textbf{371\,ms} and raises throughput to \textbf{56\,frames}.
This confirms that backlog growth behind the synchronous detector is
the main bottleneck in the baseline.

\vspace{1mm}
\noindent\textbf{Full MM-BEV.}
Adding temporal ROI and adaptive resolution (\name{}) reduces e2e
latency further to \textbf{344\,ms} (a \textbf{2.93$\times$}
reduction over baseline) and processes \textbf{60\,frames} per scene
(a \textbf{6$\times$} improvement).
The per-frame detection contribution drops from 199\,ms (baseline
model) to 95\,ms (\name{} model), consistent with the offline
benchmarks in \S\ref{sec:latency}.

\begin{table}[t]
\caption{End-to-end ROS pipeline summary.}
\label{tab:e2e}
\centering
\resizebox{\columnwidth}{!}{%
\begin{tabular}{@{}lcccc@{}}
\toprule
\textbf{Config} & \textbf{Frames/scene} &
    \textbf{E2E lat.\ mean (ms)} &
    \textbf{E2E lat.\ p95 (ms)} &
    \textbf{Speedup} \\
\midrule
BEVFusion         & 10  & 1006 & 1455 & -- \\
Scheduler Only    & 56  & 371  & 498  & $2.71\times$ \\
\textbf{MM-BEV}   & \textbf{60}  & \textbf{344}  & \textbf{434}  & $\mathbf{2.93\times}$ \\
\bottomrule
\end{tabular}}
\vspace{-2mm}
\end{table}

Table~\ref{tab:e2e} provides the full numbers.
The 95th-percentile e2e latency shrinks from 1455\,ms to 434\,ms,
confirming that the long tail of the backlog---not just the
average---is reduced.
The residual gain from ROI on top of the scheduler alone (371 to
344\,ms) reflects faster per-frame model execution translating to
a shorter queue wait time.

\subsection{Real-World Deployment}
\label{sec:deployment}

\begin{figure}[t]
\centering
\includegraphics[width=.8\columnwidth]{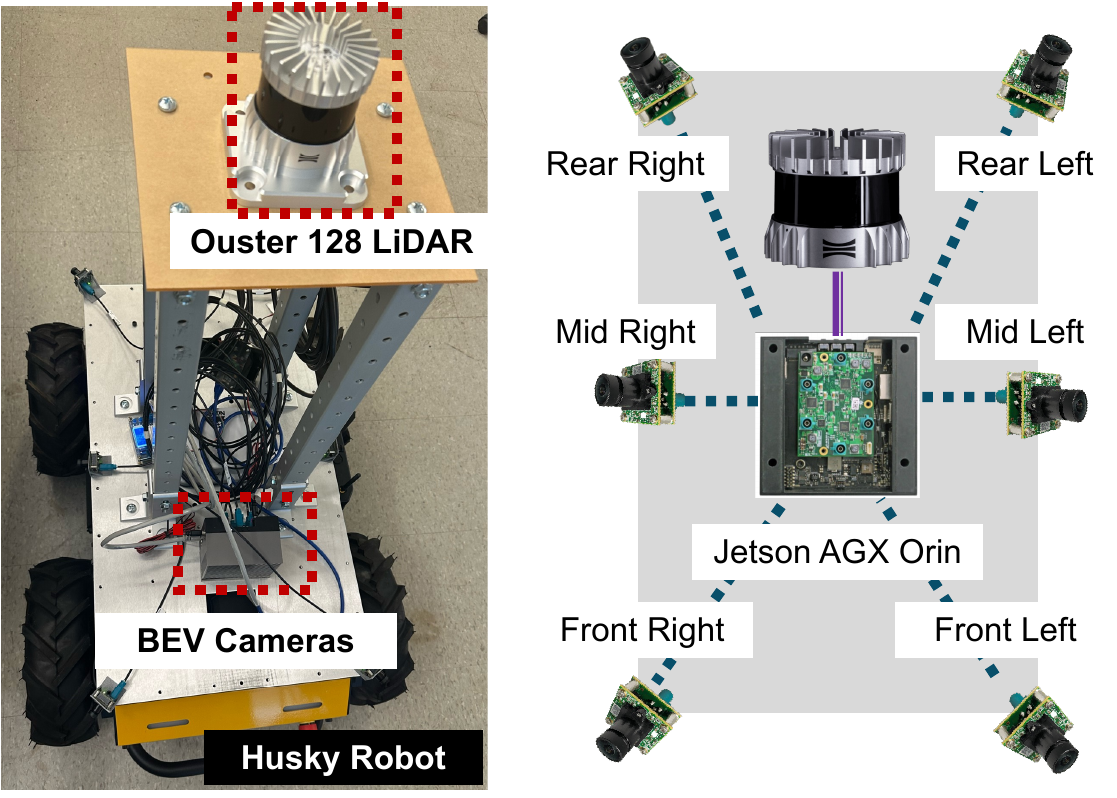}
\vspace{-2mm}
\caption{Deployment platform: Clearpath Husky\,A300 with Ouster
OS1-128 LiDAR and e-consystems AR0234 FHD GMSL2 BEV-view camera,
running \name{} on a NVIDIA Jetson AGX Orin.}
\label{fig:husky}
\vspace{-2mm}
\end{figure}



\noindent\textbf{Platform.}
We validate \name{} on a Clearpath Husky\,A300 mobile
robot~\cite{clearpath_huskya300} equipped with six e-con AR0234
GMSL2 wide-angle cameras ($1920{\times}1080$, $\sim\!86^{\circ}$
HFOV)~\cite{econ_ar0234} and an Ouster OS1-128
LiDAR~\cite{ouster_os1}; all perception runs on the onboard NVIDIA
Jetson AGX Orin with the BEVFusion backbone compiled to
TensorRT\,int8.
We collect a $22.6$-minute drive ($3{,}391$ frames) across outdoor
plazas, parking lots, and indoor corridors.
The dense baseline runs full inference every frame at a mean
$91.0$\,ms; a lightweight voxel-grid connected-component proposer
identifies object clusters in $10.5$\,ms ($21\times$ faster than
sklearn DBSCAN at $217$\,ms).

\begin{figure}[t]
\centering
\includegraphics[width=\columnwidth]{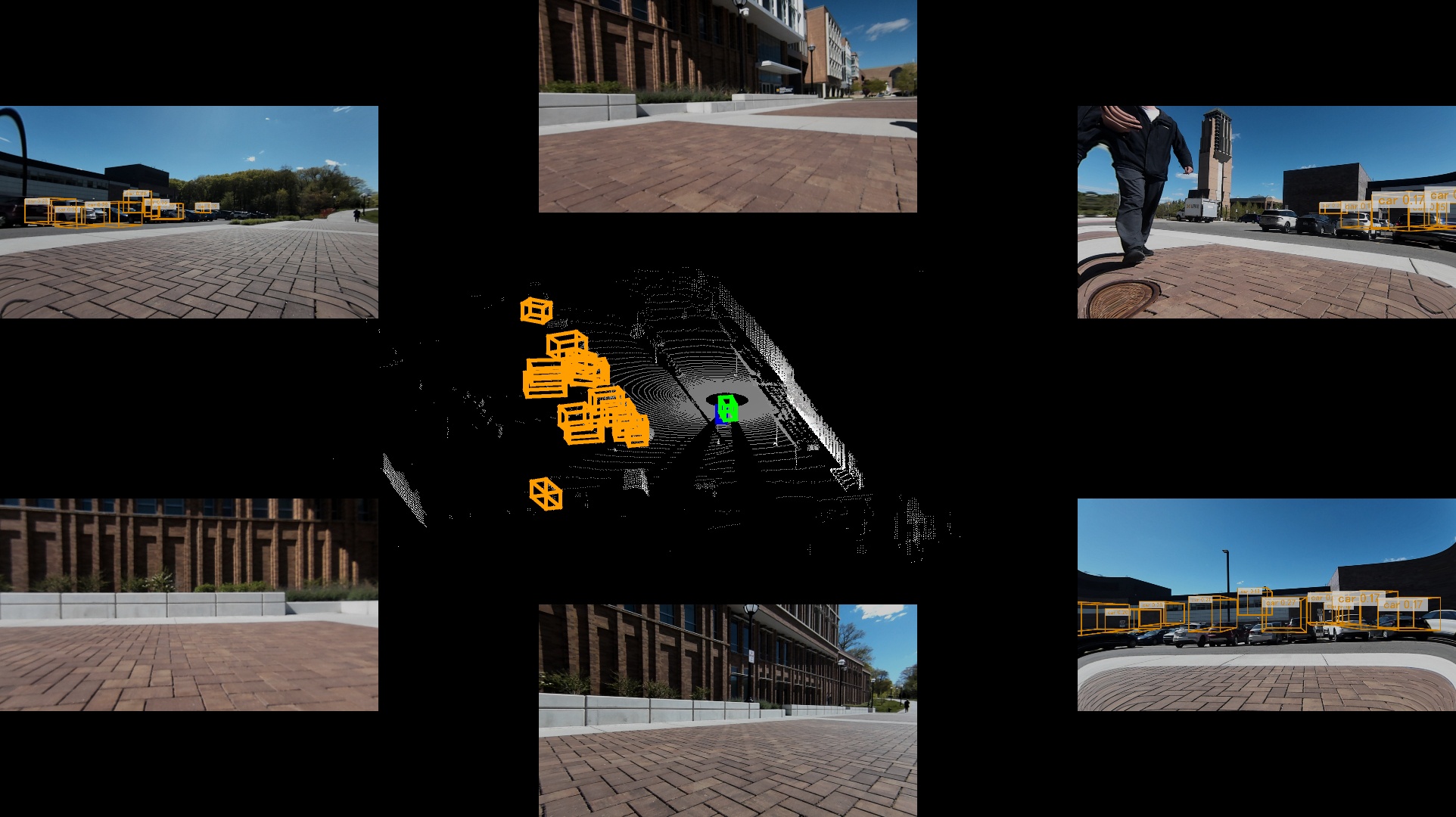}
\vspace{-5mm}
\caption{Representative deployment frame: six on-robot cameras
(surrounding tiles) and the live BEV (center) with detected objects
(orange) around the ego (green). \name{} runs on-device on the Jetson
AGX Orin and detects the row of parked vehicles and the pedestrian
crossing the robot's forward path.}
\label{fig:jetson-example}
\vspace{-2mm}
\end{figure}

\vspace{1mm}
\begin{figure}[t]
\centering
\includegraphics[width=.65\columnwidth]{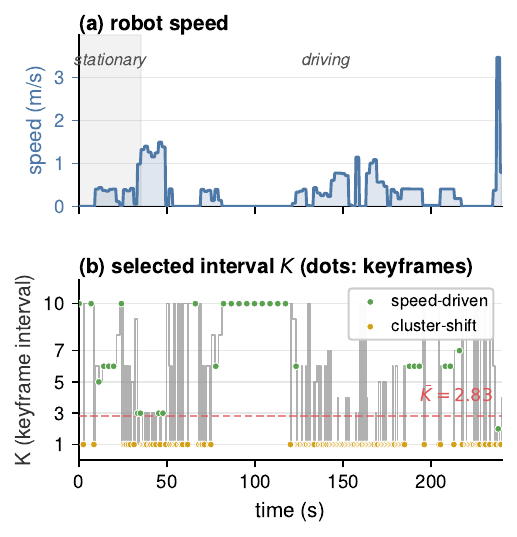}
\vspace{-5mm}
\caption{Adaptive keyframe selection over a $240$\,s drive segment.
\textbf{(a)} Robot speed from odometry.
\textbf{(b)} Selected interval $K$ (gray step); colored dots mark
emitted keyframes by trigger reason. Mean $\bar K{=}2.83$ across the
full run.}
\label{fig:jetson-schedule}
\vspace{-2mm}
\end{figure}

\vspace{1mm}
\noindent\textbf{Adaptive scheduler.}
The scheduler picks a per-frame keyframe interval
$K\!\in\![1,10]$ from the robot's odometry speed and the inter-frame
BEV cluster shift (Fig.~\ref{fig:jetson-schedule}): $K{=}10$ when
stationary (trigger ``speed''), and $K{=}1$ when the cluster set
changes between frames during active driving (trigger
``cluster-shift''). Across the drive the scheduler selects
$\bar K{=}2.83$ with $40.6\%$ of frames as keyframes, yielding a mean
per-frame inference latency of \textbf{$43.2$\,ms}---a
\textbf{$2.11\times$} reduction over the dense baseline.
By comparison, fixed $K{=}3$ and $K{=}5$ reach $32.9$\,ms
($2.77\times$) and $20.9$\,ms ($4.35\times$) but cannot react to
scene changes.
Fig.~\ref{fig:jetson-example} shows a representative frame: forward-path
vehicles and a crossing pedestrian are all detected on-device,
confirming that the temporal-ROI and adaptive-resolution gains
transfer from desktop GPU to the constrained Jetson Orin compute budget.


\section{Related Work}
\label{sec:related-work}

BEV perception has become mainstream for AVs, thanks to its comprehensive 
view understanding, which guides their planning and navigation
\cite{BEV-Perception-Survey-2024, ma2023visioncentric}. Compared to
vision-centric BEV perception that utilizes only camera inputs
\cite{li2022bevformer, yang2023bevformer, hu2023planning}, multi-modality 
BEV perception combining LiDAR and cameras is more practical and widely 
deployed, primarily due to its superior accuracy and reliability across 
diverse traffic scenarios~\cite{liang2022bevfusion,liu2023bevfusion}.

However, achieving real-time BEV perception remains challenging due mainly 
to the high computational demands from high-frequency and high-resolution 
multi-modality sensors~\cite{liu2023bevfusion}. SOTA research has primarily 
approached this challenge by either designing lightweight models or 
compressing dense models into sparse structures to improve computational 
efficiency~\cite{xie2023sparsefusion, bai2022transfusion, yan2023cross}. 
For instance, BEVFusion~\cite{liu2023bevfusion} 
utilizes dense camera and LiDAR feature maps to perform sensor fusion 
within a shared BEV space. 
Transfusion~\cite{bai2022transfusion} maintains dense features for the 
camera branch but uses sparse projection for LiDAR inputs. 
SparseFusion~\cite{xie2023sparsefusion} employs sparse features and fusion 
to reduce overhead. Cross-Modality Transformer (CMT)~\cite{yan2023cross} further explores efficient multi-modal 
integration by designing a transformer-based feature transformation.
Yet, these methods ignore the inherent sparsity in input sensor data --- not 
all pixels are equally important --- or consider redundancy only intra-modality. 
Remix and Elf recognize the variable importance of regions and but only apply
to one camera, not applied directly to multi-modality sensors~\cite{zhang2021elf, jiang2021flexible}. FLEX considers BEV cameras but only focuses on frame-level ROIs~\cite{xu2024flex}.
MMEdge~\cite{huang2026mmedge} pipelines on-device multimodal inference
by overlapping sensing and encoding into fine-grained units with
cross-modal speculative skipping, but operates on classification-style
fusion tasks rather than dense BEV detection.
Another challenge is cross-modal
synchronization: LiDAR and cameras operate at different frequencies, and poor
alignment of their data can degrade performance. RT-BEV~\cite{liu2024rt}
tackles the ROI and synchronization issues of cameras only. \name{} extends it to
a more practical LiDAR--camera combination, leveraging both input sparsity
and robust alignment for efficient BEV perception.

\section{Conclusion}
We presented \name{}, a real-time multi-modality BEV perception system
organized around the principle of computing \emph{where and when it
matters}. \name{} decomposes BEV detection into a \emph{mandatory}
part---the critical objects inside the ego's braking distance and
short time-to-collision---and an \emph{optional} part, and combines
four mechanisms (a criticality-ranked temporal-ROI selector, sparse
ROI-aware feature extraction, a latency-aware coordinator, and an
asynchronous streaming scheduler).
On nuScenes---where only $11\%$ of annotated objects are mandatory
under the criticality model---\name{} cuts inference latency by
$1.96\times$ and end-to-end latency by $2.93\times$, preserves
geometry-critical recall exactly, and drops safety-critical recall by
only $0.2$ percentage points. On a Clearpath Husky\,A300 robot with Jetson AGX Orin, \name{} reaches a further $2.11\times$ mean-latency
reduction under an adaptive keyframe schedule, confirming that the
gains transfer to embedded hardware. These results show that
criticality-aware scheduling is a practical foundation for real-time
multi-modality perception on resource-limited AVs.


\newpage
\bibliographystyle{IEEEtran}
\bibliography{main}

\end{document}